\pdfoutput=1
\documentclass[11pt]{article}

\usepackage[final]{acl}

\usepackage{times}
\usepackage{latexsym}

\usepackage[T1]{fontenc}
\usepackage[utf8]{inputenc}

\usepackage{microtype}

\usepackage{inconsolata}

\usepackage{graphicx}

\usepackage{times}
\usepackage{enumitem}
\setlist[itemize,enumerate]{topsep=4pt,itemsep=0pt,leftmargin=*}

\usepackage[utf8]{inputenc} % allow utf-8 input
\usepackage[T1]{fontenc}    % use 8-bit T1 fonts
\usepackage{hyperref}       % hyperlinks
\usepackage{url}            % simple URL typesetting
\usepackage{amsfonts}       % blackboard math symbols
\usepackage{nicefrac}       % compact symbols for 1/2, etc.
\usepackage{microtype}      % microtypography
\usepackage{xcolor}         % colors
\usepackage{booktabs}
\usepackage{amsmath}
\usepackage{multirow}
\usepackage{graphicx}
\usepackage{enumitem}
\usepackage{subcaption}
\usepackage{caption}
\usepackage{rotating}
\usepackage[normalem]{ulem}
\usepackage{amssymb}
\usepackage{colortbl}
\usepackage{linguex}

\usepackage{tikz,pgfplots,pgfplotstable}

\usepackage{siunitx} 
\usepackage{multirow, makecell}
\usepackage{pifont}% http://ctan.org/pkg/pifont
\usepackage{siunitx} 
\usepackage{multirow, makecell}
\usepackage{tabularx}% added for table design
\usepackage[all]{nowidow}
\definecolor{darkgreen}{rgb}{0.0, 0.42, 0.24}
\definecolor{green}{RGB}{112, 173,71}
\definecolor{blue}{RGB}{68, 114,196}
\definecolor{orange}{RGB}{237, 125,49}
\definecolor{red}{RGB}{202, 54,49}
\definecolor{yellow}{RGB}{222,194, 142}
\usepackage{xspace}
\usepackage{arydshln}

\newcommand{\google}{\textsc{GoogleTranslate}\xspace}
\newcommand{\towersmall}{\textsc{TowerInstruct}-\textsc{7b}\xspace}
\newcommand{\almar}{\textsc{ALMA}-\textsc{13b}-R\xspace}
\newcommand{\alma}{\textsc{ALMA}-\textsc{13b}\xspace}
\newcommand{\nllb}{\textsc{NLLB}-\textsc{54b}\xspace}
\newcommand{\towerlarge}{\textsc{TowerInstruct}-\textsc{13b}\xspace}
\newcommand{\gpt}{\textsc{GPT}-4\xspace}
\newcommand{\toweraligndata}{\textsc{MT-Pref}\xspace}

\usepackage{xcolor}

\newcommand{\cmss}[1]{{\fontfamily{cmss}\selectfont{#1}}}

\definecolor{darkblue}{rgb}{0,0,.5}
\definecolor{darkgreen}{rgb}{0,.5,0}
\definecolor{lightgray}{rgb}{.8,.8,.8}
\definecolor{aliceblue}{rgb}{0.75, 0.75, 1.0}
\definecolor{darkseagreen}{rgb}{0.46, 0.74, 0.46}
\definecolor{alizarin}{rgb}{0.82, 0.1, 0.26}
\definecolor{airforceblue}{rgb}{0.36, 0.54, 0.66}
\definecolor{red_graph}{rgb}{0.98, 0.8, 0.8}
\definecolor{blue_graph}{rgb}{0.8, 0.98, 0.8}
\definecolor{red}{rgb}{0.8, 0.0, 0.0}

\usepackage{pgfplots}
\pgfplotsset{compat=1.17}
\usepackage{pgfplotstable}

\definecolor{CustomBlue}{RGB}{57,83,191}
\definecolor{CustomRed}{HTML}{a75151}
\definecolor{DarkGreenOne}{RGB}{106,168,79}
\usepackage[most]{tcolorbox}

\newtcbox{\clustertab}[1]{on line, box align=base, colback={#1},colframe={#1},size=fbox,arc=2pt,top=-1.5pt, bottom=-1.5pt, left=-1.5pt, right=-1.5pt, boxrule=0pt, enlarge left by=1pt}

\newcommand{\firstcluster}{{\footnotesize\clustertab{orange!75}{1}}}
\newcommand{\secondcluster}{{\footnotesize\clustertab{orange!60}{2}}}
\newcommand{\thirdcluster}{{\footnotesize\clustertab{orange!50}{3}}}
\newcommand{\fourthcluster}{{\footnotesize\clustertab{orange!30}{4}}}
\newcommand{\fifthcluster}
{{\footnotesize\clustertab{orange!15}{5}}}
\newcommand{\sixthcluster}
{{\footnotesize\clustertab{orange!10}{6}}}
\newcommand{\seventhcluster}
{{\footnotesize\clustertab{orange!5}{7}}}

\title{Modeling User Preferences with Automatic Metrics: \\ Creating a High-Quality Preference Dataset for Machine Translation}

\author{Sweta Agrawal$^{1}$, José G. C. de Souza$^{3}$, Ricardo Rei$^{3}$, António Farinhas$^{1,2}$,\\  \textbf{Gonçalo Faria}$^{1}$,  \textbf{Patrick Fernandes}$^{1,2, 5}$,  \textbf{Nuno M Guerreiro}$^{1,2,3,6}$, \textbf{André F.T. Martins}$^{1,2,3,4}$\\ 
$^1$Instituto de Telecomunicações, $^2$Instituto Superior Técnico, Universidade de Lisboa\\
$^3$Unbabel, $^4$ELLIS Unit Lisbon, $^5$Carnegie Mellon University\\
$^{6}$MICS, CentraleSupélec, Université Paris-Saclay, France\\
\texttt{swetaagrawal20@gmail.com}}

\begin{document}
\maketitle
\begin{abstract} 
Alignment with human preferences is an important step in developing accurate and safe large language models. This is no exception in machine translation (MT), where better handling of language nuances and context-specific variations leads to improved quality. 
However, preference data based on human feedback can be very expensive to obtain and curate at a large scale. 
Automatic metrics, on the other hand, can induce preferences, but they might not match human expectations perfectly. 
In this paper, we propose an approach that 
leverages the best of both worlds. 
We first collect sentence-level quality assessments from professional linguists on translations generated by multiple high-quality MT systems and evaluate the ability of current automatic metrics to recover these preferences. 
We then use this analysis to curate a new dataset, \toweraligndata (\textbf{\textsc{M}}etric-induced \textbf{\textsc{T}}ranslation \textbf{\textsc{Pref}}erence), 
which comprises 18k instances covering 18 language directions, using texts sourced from multiple domains post-2022. We show that aligning \textsc{Tower} models on  \toweraligndata significantly improves translation quality on WMT23 and FLORES benchmarks.\footnote{We release the code and datasets to reproduce all the results at \url{https://github.com/deep-spin/mt-pref-alignment}.} 

\end{abstract}

\section{Introduction}

% \andre{this intro is quite long and verbose. I trimmed a bit, but we should probably trim more} \sa{trimmed a bit more..}
The use of large language models (LLMs) in machine translation (MT) has garnered significant attention from the research community \citep{kocmi-etal-2023-findings}. Unlike traditional sequence-to-sequence MT models trained on parallel data \cite{koehn-knowles-2017-six}, LLM-based MT systems either use in-context learning to elicit translation knowledge acquired during pre-training \cite{briakou-etal-2023-searching} or undergo supervised finetuning (SFT) on high-quality translations to further enhance their translation capabilities \cite{li2024eliciting, xu2023paradigm, alves-etal-2023-steering, alves2024tower}. 

The default SFT approach for LLM-based MT is to tune systems based solely on \textit{single} human reference translations. 
However, this kind of supervision might be insufficient to push quality further: First, because \textit{many} valid translations may exist for a given source, with some \textit{preferred} over others \cite{mayhew-etal-2020-simultaneous}. 
Second, because the next-token prediction objective of SFT does not capture sentence-level semantics and quality criteria \cite{eikema-aziz-2020-map, liu-etal-2022-brio}. 
This has motivated new approaches that go beyond SFT to leverage translation preferences or quality feedback to improve learning \citep{yang2023direct, he2024improving, xu2024contrastive, Zhu2024}. %\citep{shen-etal-2016-minimum,kreutzer-etal-2018-neural,yang2023direct, he2024improving, xu2024contrastive, Zhu2024}. 

%As the translation quality of LLMs continues to improve, tuning these systems based solely on \textit{single} human reference translations, as typically done in SFT, may be insufficient feedback to yield further quality gains \cite{xu2024contrastive, Zhu2024}. This is because many valid translations exist for a given source, with some \textit{preferred} over others \cite{mayhew-etal-2020-simultaneous}.
%Furthermore, the next-token prediction objective used in SFT does not capture sentence-level semantics and quality criteria \cite{eikema-aziz-2020-map, liu-etal-2022-brio}. 
%Hence, new approaches have been proposed that go beyond SFT, and instead optimize objectives that use translation preferences or quality feedback to improve learning \citep{yang2023direct, he2024improving, xu2024contrastive, Zhu2024}. 
% \patrick{Given the current size of the intro, this paragraph might be a bit long. If we need to cut, I would probably start here}\sa{will do, once everything is in, I am still waiting for 1-2 results and analysis.}

A key factor in aligning LLMs toward translation preferences is 
ensuring the quality and diversity of the datasets used for training \cite{gao2024impact, morimura2024filtered, liu2023makes}. Unfortunately, existing datasets have several limitations: First, they are created from translation outputs of one or two models, for limited language pairs, 
thereby restricting their diversity and applicability to novel scenarios.
Second, these datasets are either entirely automatically generated \cite{xu2024contrastive} or completely based on human feedback \cite{Zhu2024}. While automatic evaluation offers efficiency, it lacks the crucial validation that the metrics used truly align with human preferences.  On the other hand, datasets that use human feedback, while high-quality and reliable, pose resource constraints and are challenging to scale. 

% \patrick{small nitpick, but on first reading this paragraph (and something similar on the abstract), I was "lead" into believing that the novelty would be a preference dataset mixing human preferences and metric-preferences, while its more like using human preferences to inform the construction of preference dataset using automatic metrics right? } \sa{yes, do you think this needs to be framed differently?} \patrick{potentially: a simple softening would be "balances the advantages of automated metrics while ensuring they lead to preferences that aligns with humans.} \patrick{this also triggers another question? could we compare with a baseline that uses the preferences from the human annotations we collected} \sa{the preferences were on WMT23 and that's the main evaluation set, so unfortunately no :( but maybe I can do 1 exp and check on flores?} \nuno{I agree with Pat on the "first reading" thing, and like his suggestion} \sa{Already rephrased and also running human pref baselines} 

To bridge this gap, we provide a holistic approach to balance the advantages of automated metrics while ensuring that they lead to preferences that truly align with humans. 
We first collect sentence-level quality assessments and preferences from human expert translators (\S\ref{sec:human_metric})---we use the WMT23 English-German and Chinese-English datasets \cite{kocmi-etal-2023-findings} with outputs from five high-quality MT systems: \towersmall, \towerlarge \cite{alves2024tower}; \textsc{ALMA-13b-R} \cite{xu2024contrastive}; GPT-4-based \cite{hendy2023good} and \textsc{GoogleTranslate}. 
Using these assessments, we then examine the ability of automatic quality estimation (QE) metrics to recover human preferences.  %(\S\ref{sec:eval_qe}). 
Our findings show that an ensemble of \textsc{xComet-xl} and \textsc{xComet-xxl} \cite{guerreiro2023xcomet}---\textsc{xComet-xl+xxl}---achieves the highest correlation with human judgments and a high precision score in identifying the preferred translations.

Using this analysis, we create a new MT preference dataset, \textit{\toweraligndata} (\textbf{\textsc{M}}etric-induced \textbf{\textsc{T}}ranslation \textbf{\textsc{Pref}}erence dataset), with source sentences mined post-2022 for 10 languages (English, German, Chinese, Russian, Portuguese, Italian, French, Spanish, Korean and Dutch). Translations for each source sentence are generated using diverse MT systems representing different architectures, training data, and quality levels (\S\ref{sec:data}).  %(\S\ref{subsec:dataandmodel}). 
We use the ensemble metric \textsc{xComet-xl+xxl} to get the most and least preferred translations from the set of hypotheses. Experiments on aligning MT-specialized decoder-only models (\textsc{Tower}) using existing preference learning algorithms with our \toweraligndata dataset demonstrate improved translation quality on the WMT23 \cite{kocmi-etal-2023-findings} and FLORES \cite{costa2022no} benchmarks, with larger gains in out-of-English translation directions (\S\ref{sec:results}). 
Further analysis shows that the aligned models better rank translations according to human preferences over baselines.

\section{Background: Aligning MT with Translation Preferences} \label{sec:pref_methods}

Given a source text, the goal of MT is to generate a translation that accurately reflects the information and meaning conveyed in the source. %An MT model parameterized by $\theta$, is trained to maximize the log-likelihood of a reference translation $y$, given an input $x$ encoding the text to be translated and optionally a translation instruction (in case of an LLM) as:
At training time, the MT model $\pi_\theta$ goes through SFT to minimize the negative log-likelihood (NLL) loss induced by source-reference pairs $(x, y)$: % $\{(x^{(i)}, y^{(i)})\}_{i=1}^N$:
\begin{equation}
    \mathcal{L}_{\text{NLL}} (x, y; \theta) = -\log \pi_\theta \left(y | x \right).
\end{equation}
%\begin{equation}
%    \mathcal{L}_{\text{NLL}} (\theta) = -\sum_{i=1}^N \log p \left(y^{(i)} | x^{(i)}; \theta \right).
%\end{equation}
A drawback of SFT is that it typically 
optimizes the model towards a \textit{single} reference translation. 
In contrast, preference learning objectives incorporate relative preferences between alternatives, allowing the model to learn from subtle differences in translation quality %, even when these alternatives are not strictly valid or correct 
\citep{zeng2023tim}. %The important requirement being that the alternative options capture meaningful differences in translation quality.
% \gf{It highlights the power of pretraining and its relative importance over large-scale instruction tuning and reinforcement learning approaches (cite lima paper). Giving alignment phases more of a role of expose the knowledge and capabilities that were already acquired during pretraining than instill new ones. } \sa{need to incorporate this, I restructured things too much, hence its not clear where this was meant to go originallu}

Different variants of preference optimization (PO) have been proposed in the literature. Reinforcement learning from human feedback (RLHF) has shown to be effective in aligning model behavior with human values \cite{ouyang2022training}. 
% Typically, this requires collecting human preferences or feedback on multiple outputs generated by language models given task instructions; fitting a reward model on those preferences, and using reinforcement learning (RL) methods to incorporate this reward model in the training objective. 
\citet{rafailov2024direct} propose direct preference optimization (DPO) as a simple and scalable alternative to RLHF.  %using a reformulation of the KL-constrained reward maximization problem under the Bradley-Terry model of preferences. 
Given a preference dataset $\mathcal{D}$ with  
source sentences $x$, preferred or chosen outputs $y_+$ and less preferred or rejected outputs $y_-$, the model is trained with the following objective:
%$\mathcal{D} = \{(x^{(i)}, y_{+}^{(i)}, y_{-}^{(i)})\}_{i=1}^{N}$  with source sentences $x^{(i)}$, preferred or chosen outputs $y_+^{(i)}$ and less preferred or rejected outputs $y_-^{(i)}$, the model is trained with the following objective:
%\begin{multline} \label{eq:dpo}
%\small
%     \mathcal{L}_{\text{DPO}} (\pi_{\theta}; \pi_\text{ref}) = -\mathbb{E}_{(x, y_+, y_-)\sim \mathcal{D}} \\ \left[ \log \sigma \left( \beta \log \frac{\pi_{\theta}(y_+ | x)}{\pi_\text{ref}(y_+ | x)} 
%   -  \beta \log \frac{\pi_{\theta}(y_- | x)}{\pi_\text{ref}(y_- | x)} \right) \right] ,
%\end{multline}
\begin{align} \label{eq:dpo}
\small
     &\mathcal{L}_{\text{DPO}} (x, y_\pm; \pi_{\theta}, \pi_\text{ref}) = \\
     &-\log \sigma \left( \beta \log \frac{\pi_{\theta}(y_+ | x)}{\pi_\text{ref}(y_+ | x)} 
   -  \beta \log \frac{\pi_{\theta}(y_- | x)}{\pi_\text{ref}(y_- | x)} \right),\nonumber
\end{align}
where $\pi_\theta$ is the parameterized policy, 
$\pi_\text{ref}$ is a base reference policy (set to the policy used to generate the dataset for collecting preferences), % when available or initialized with $\pi_\text{ref}=\arg\max_{\pi} \mathbb{E}_{x,y_+ \sim D} [\log \pi(y_+ | x)]$ to mitigate the distribution shift between the preference dataset and the base model's policy
and 
$\beta$ is a (inverse) temperature hyperparameter. % controlling the deviation from a base reference policy $\pi_\text{ref}$.  $\pi_\text{ref}$ is set to the policy used to generate the dataset for collecting preferences when available or initialized with $\pi_\text{ref}=\arg\max_{\pi} \mathbb{E}_{x,y_+ \sim D} [\log \pi(y_+ | x)]$ to mitigate the distribution shift between the preference dataset and the base model's policy. 

One notable limitation of the DPO
% $\mathcal{L}_\text{DPO}$ \antonio{DPO instead of $\mathcal{L}_\text{DPO}$? to be consistent with the others} 
objective is that it requires both $\pi_{\theta}$ and $\pi_\text{ref}$ in memory, significantly increasing memory requirements and computation costs. 
To address this, \citet{xu2024contrastive} further approximate the DPO objective using a uniform reference model ($\pi_\text{ref} = \mathcal{U}$) to derive a contrastive preference optimization (CPO) loss: 
% \andre{introduce the acro CPO? use $\mathcal{L}_{\text{CPO}}$ instead of  $\mathcal{L}_{\text{prefer}}$?} \sa{they call eq~\ref{eq:cpo} the CPO loss }
\begin{align} \label{eq:prefer}
     &\mathcal{L}_{\text{CPO}} (x, y_\pm; \pi_{\theta}, \mathcal{U}) =  -\log \sigma \left( \beta \log \frac{\pi_{\theta}(y_+ | x)}{\pi_{\theta}(y_- | x)} \right).
\end{align}
%\begin{multline} \label{eq:prefer}
%   \mathcal{L}_{\text{prefer}} (\pi_{\theta}; \mathcal{U}) = -\mathbb{E}_{(x, y_+, y_-)\sim D} \\ \left[ \log \sigma \left( \beta \log \pi_{\theta}(y_+ | x) - \beta \log \pi_{\theta}(y_- | x)  \right) \right].
%\end{multline}
% \noindent This increases operational efficiency as CPO only requires the parameterized model in the memory, enabling the training of larger models at reduced costs compared to DPO while maintaining or improving translation quality. 
However, both losses \eqref{eq:dpo}--\eqref{eq:prefer} only maximize the relative difference between preferred and dispreferred outputs. On tasks like MT where the difference in the two outputs is small, this may lead to failure modes where the learning objective leads to a reduction of the model’s likelihood of the preferred examples, as long as the relative probability between the two classes increases \cite{pal2024smaug}. 
Therefore, following \citet{hejna2023contrastive}, \citet{xu2024contrastive} introduce a behavior cloning regularizer to ensure that the model stays close to the preferred distribution, %\textit{i.e.}, assigns lower loss when the policy has higher likelihood of preferred translations. This results in the following 
leading to the final CPO objective: 
% \andre{is this relevant? are we using this? should we give a name to this loss?}\sa{This is the CPO objective}
\begin{align} \label{eq:cpo}
\small
&\mathcal{L}_\text{CPO} (x, y_\pm; \theta) = \\ & \quad \mathcal{L}_{\text{DPO}} (x, y_\pm; \pi_\theta, \mathcal{U}) +  \lambda \mathcal{L}_{\text{NLL}}  (x, y_+; \theta), \nonumber
\end{align}
where $\lambda$ is a hyperparameter that controls the relative strength of the two objectives.

As the quality of the preference datasets used for training is key for its success \cite{gao2024impact, morimura2024filtered, liu2023makes}, we next discuss our process of collecting a high-quality dataset for preference learning for MT.

\section{Modeling User Preferences Via Automatic Metrics} \label{sec:human_metric}

To create a high-quality preference dataset for MT, we need human judgments on translation outputs from strong MT systems.
This helps us understand and model human preferences among competitive translations.
Since large-scale collection of these judgments is costly, we evaluate existing automatic metrics to see if they effectively reflect human preferences.
This determines if metrics can be reliable proxies for human judgments when translation quality is high and preference variance is low.

We describe the dataset, models, and task instructions given to the expert annotators used in our study in \S\ref{sec:task}. The human evaluation results are presented in \S\ref{sec:findings}. Finally, we discuss %the findings from 
our meta-evaluation of automatic MT metrics in their ability to recover human preferences 
in \S\ref{sec:eval_qe}.

\subsection{Data and Annotation Task} \label{sec:task}

%\andre{how many? do we compute IAA? etc.}\sa{one linguist per language, also added the time to do assessment} 
We randomly sample 200 source instances from the WMT23 English-German (\textsc{en-de}) and Chinese-English (\textsc{zh-en}) test sets and 
generate translations using five MT models: \google, \gpt, \towersmall, \towerlarge, and \almar (described in Appendix~\ref{sec:model_des}).\footnote{This is the only dataset that was not used in the training of any evaluated models.}
We employ DA+SQM (Direct Assessment + Scalar
Quality Metric) source contrastive evaluation \cite{kocmi-etal-2022-findings}, using the Appraise evaluation framework \cite{federmann-2018-appraise}.\footnote{\url{https://github.com/AppraiseDev/Appraise.git}.}
We then ask one linguist per language pair to read all translations for a given source and evaluate each of them on a continuous 0-100 scale. The scale features seven labeled tick marks indicating different quality labels combining \textit{accuracy} and \textit{grammatical correctness}. Linguists can further adjust their ratings to reflect preferences or assign the same score to translations of similar quality.
Detailed guidelines, compensation details, and a screenshot of the interface are provided in Appendix~\ref{sec:guidelines}.
This results in a preference dataset including 1000 ratings each for \textsc{en-de} and \textsc{zh-en}.\footnote{Completing the task takes approximately 10 to 11 hours for each language pair.}

\subsection{Human Evaluation Findings}  \label{sec:findings}

We present the results from our human evaluation in Table~\ref{tab:human_eval} and discuss the findings below:

\begin{table}[!t]
    \centering
    \setlength{\tabcolsep}{2pt}
    % \scalebox{0.85}{
    \resizebox{\linewidth}{!}{
    \begin{tabular}{lccccc}
    \toprule
     & \multicolumn{2}{c}{\textbf{\textsc{DA}}} &&  \multicolumn{2}{c}{\textbf{\textsc{Top-1}}}\\
     \cline{2-3} \cline{5-6}
     \addlinespace[0.1cm]
   \multicolumn{-2}{l}{\textbf{\textsc{MODEL}}} & \textbf{\textsc{EN-DE}} & \textbf{\textsc{ZH-EN}}  && \textbf{\textsc{EN-DE}} & \textbf{\textsc{ZH-EN}}  \\
   \midrule
    \addlinespace[0.5em]
   \google & 86.87 & 79.85 && 62 & 114 \\
    \gpt & 87.98 & 79.12  && 66 & 108 \\ 
 \towerlarge & 86.53 & 69.12 && 53 & 56  \\
 \almar & 84.96 & 66.02  && 46 & 51  \\
  \towersmall & 83.32 & 68.66 && 37 & 63 \\
  \bottomrule
    \end{tabular}
    }
    \caption{Human evaluation results: DA scores for all MT systems are high, suggesting that translations are generally of very good quality according to experts. 
    }
    \label{tab:human_eval}
\end{table}

% for placing

\begin{table*}[!t]
    \centering
    \resizebox{\linewidth}{!}{
    \begin{tabular}{lrrrccrrrc}
    \toprule
     & \multicolumn{4}{c}{\textbf{EN-DE}} &&  \multicolumn{4}{c}{\textbf{\textsc{ZH-EN}}}\\ 
     \cline{2-5}  \cline{7-10}
     \addlinespace[0.1cm]
  \multirow{-2}{*}{\textbf{\textsc{metric}}}  & \textsc{P} & \textsc{S} & \textsc{Tau}  & \textsc{Precision@1}   && \textsc{P} & \textsc{S} & \textsc{Tau}    & \textsc{Precision@1}   \\ 
  \midrule
\textsc{CometKiwi}-\textsc{xl} &
0.275 & 0.272 & 0.229 & \textbf{47.0} &&
0.332 &0.336 &0.289   & 42.9 \\

\textsc{CometKiwi}-\textsc{xxl} &
0.253 & 0.238 & 0.198 & 43.9&&
0.342 &0.346&0.279 & 46.6  \\

 \addlinespace[0.05cm]
\textsc{xComet}-\textsc{xl} &
{0.334} &0.300 &0.249 & 41.9    &&
\textbf{0.456} &{0.410} &\textbf{0.342}& 44.5  \\
\textsc{xComet}-\textsc{xxl} &0.316 & 
{0.312} & {0.252} & 44.4  &&
0.343 &{0.410} &0.340  & 44.0   \\

 \addlinespace[0.05cm]
\textsc{MetricX-23}-\textsc{l} & 
0.238 &0.238 &0.191   & 37.9  &&
0.428 &0.409 &0.328& 42.4 \\
\textsc{MetricX-23}-\textsc{xl} &
0.270 &0.245 &0.206   & 39.4  &&
0.417 &{0.410} &\textbf{0.342}  & 45.5\\
%  \addlinespace[0.1cm]
% \gpt & 0.321 & 0296 & 0.250 &  &  0.449 & \textbf{0.436} & \textbf{0.369} &  \\
\addlinespace[0.1cm]
\hdashline
\addlinespace[0.1cm]

\textsc{xComet-xl+xxl} &
\textbf{0.341} &\textbf{0.329} &\textbf{0.270}  & \textbf{47.0}   && 
{0.434} &\textbf{0.411} &0.336  & \textbf{48.7} \\
\textsc{CometKiwi-xl+xxl} & 0.273 & 0.252 & 0.211& 43.9 &&  
0.347 & 0.357 & 0.290 & 41.4\\
\textsc{xComet+Kiwi}-\textsc{xxl} &
0.286 &0.271 &0.223 & 45.5   &&
0.377 &0.382 &0.304  & 46.6  \\

\addlinespace[0.1cm]
\hdashline
\addlinespace[0.1cm]
\textsc{COMET-ref} & 0.331 & 0.286 & 0.234 & 50.5  &&
 0.243 & 0.211 & 0.169 & 47.1 \\
\bottomrule
    \end{tabular}
    }
    \caption{Correlation and Precision@1 for automatic QE metrics: \textsc{xComet-xl+xxl} results in the highest correlation and Precision@1 across the board, outperforming reference-based metric, COMET-\textsc{ref}.
    % \andre{it's nicer if the numbers are in text mode as I did in the previous table. also, it would be good to try to avoid the negative vertical spacing if we can. for example you can remove the row which says ``Ensembles'' and eats a lot of space, add a horizontal line and explain in the caption that that section are ensembles }
    }
    \label{tab:automaticmt}
    % \vspace{-0.5cm}
\end{table*}

% \begin{figure*}[t]
%  \centering
% \begin{subfigure}{0.47\textwidth}
%   \centering
%   \includegraphics[width=\linewidth]{figs/ende-pair-human.pdf}
% \end{subfigure}%
% % \hspace{0.3cm}
% \begin{subfigure}{0.47\textwidth}
%   \centering
%   \includegraphics[width=\linewidth]{figs/zhen-pair-human.pdf}
% \end{subfigure}% 
%     \caption{ Pairwise Preferences for different Systems on English-German (left) and Chinese-English (right) datasets.  
%     % \nuno{any reason to use absolute counts instead of relative frequency? it could make for better reading. Design-wise: these colours feel like they are screaming at me haha; try to be consistent in the colour palettes throughout the paper~(Figure 2, for example, uses a different palette) and the tables are quite neat with the grays.} 
%     % \todo{relative frequency (suggested by nuno) and make it neater}
%     }
%       \label{fig:pairwise}
% \end{figure*}

\begin{figure}[t]
 \centering
\begin{subfigure}{0.45\textwidth}
  \centering
  \includegraphics[width=\linewidth]{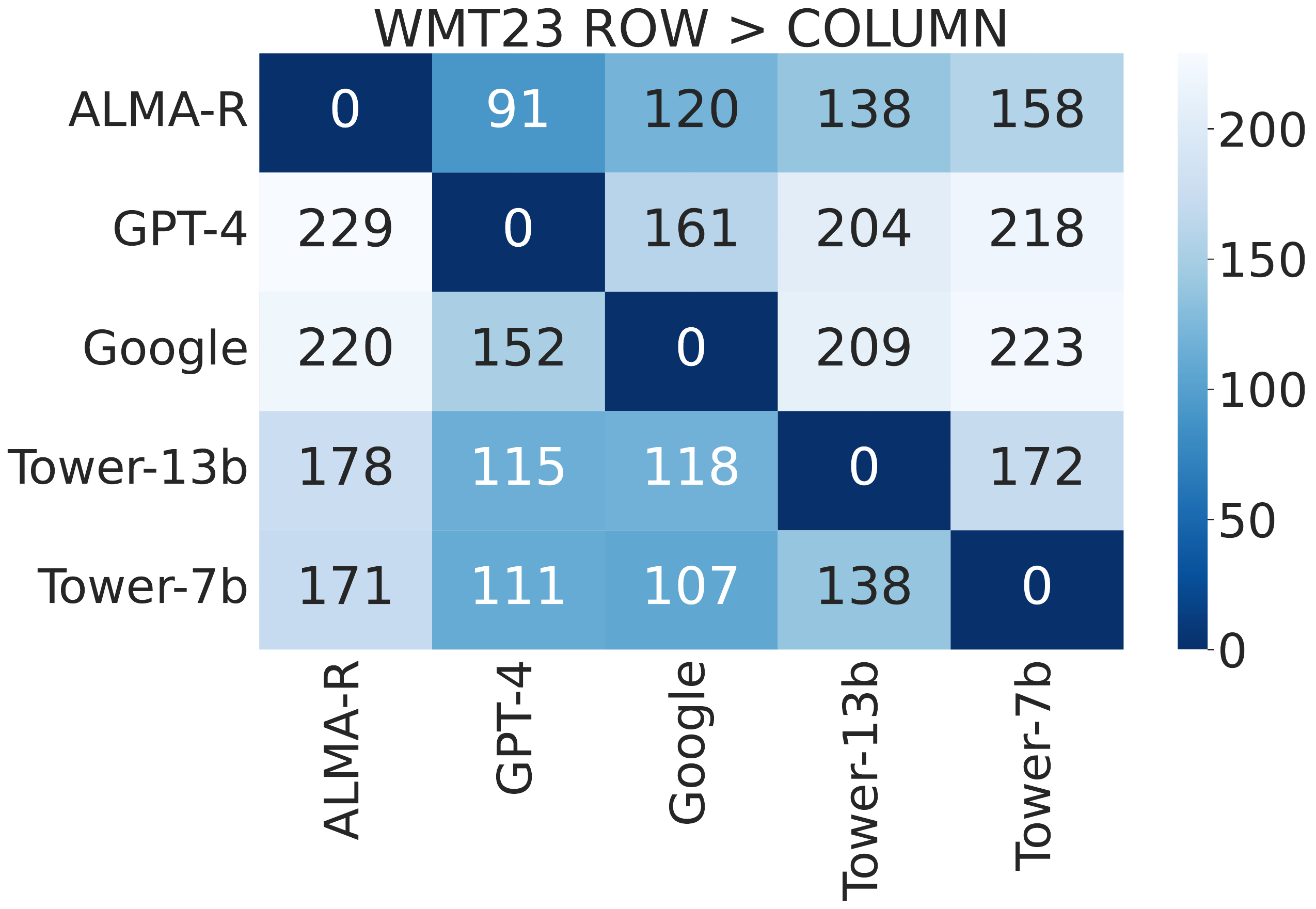}
\end{subfigure}
    \caption{ Pairwise Preferences between different Systems: Google and GPT-4 translations are more preferred over open-sourced alternatives.  
    }
      \label{fig:pairwise}
\end{figure}

\paragraph{Overall Quality} For \textsc{en-de}, DA scores range from 83.32 to 87.98, with no significant difference in the translation quality of different systems, according to the Mann-Whitney test \cite{mcknight2010mann}. On the other hand, DA and Top-1 are significantly better for \gpt and \google models for \textsc{zh-en}. Further qualitative analysis shows that for WMT23 \textsc{zh-en}, the quality of the source sentences is often poor---up to 25\% of source sentences were marked as problematic by the linguist. This suggests there is still room for improvement for open-source models over close-sourced alternatives when generating translations for noisy source texts \cite{peters2024did}. 

\paragraph{Pairwise Preferences} We also report pairwise wins for each model against the other in Fig.~\ref{fig:pairwise}. \google and \gpt outputs are generally more preferred over open-sourced translation alternatives. Further analysis shows that about 25\% and 10\% pairs are tied for equal preferences for \textsc{zh-en} and \textsc{en-de} respectively, further validating close translation quality amongst alternatives. 
Taken together, these results show that all the evaluated MT systems generate high-quality translations. %, with a slight preference for closed-source models over open alternatives.

\subsection{Evaluating Automatic Metrics}  \label{sec:eval_qe}

% We now turn towards assessing whether automatic quality estimation metrics can recover the human preference rankings on this dataset. This allows us to gauge whether current metrics can serve as proxies for human judgments when the translation quality is high and variance in preferences is low. 

% \nuno{I would remind the reader here why this step is relevant; they have read about it in the introduction, but, by now, they have been through lots of other content.} \sa{i added the second sentence but maybe we can elaborate more?}

We evaluate the best-performing metrics from the WMT23 QE Shared Task: 1) \textsc{CometKiwi}  \cite{rei-etal-2023-scaling}; 2) \textsc{xComet} \cite{guerreiro2023xcomet}; 3) \textsc{MetricX} \cite{juraska-etal-2023-metricx} and ensembles of these metrics obtained by averaging the scores from the two metrics: 4) \textsc{CometKiwi-xl+xxl} 5) \textsc{xComet-xl+xxl} and 6) \textsc{CometKiwi+xComet-xxl}.\footnote{We refer the reader to the original papers for each metric for more details about the training and architecture.}

\subsubsection{Metrics for Meta-Evaluation}

We report the following scores to assess these metrics in their ability to recover human preferences at the segment level: 

\paragraph{Correlation} Following WMT evaluation campaign, we report the Pearson (P), Spearman (S), and Kendall Tau (\textsc{tau}) correlation of automatic metrics with human judgements over all collected judgments grouped by source.

\paragraph{Precision@1 for the best translation} We additionally report the precision of identifying the best hypothesis by an automatic metric as the number of times the metric's ranked best translation is in the set of human-ranked best translations. Note that as we ask linguists to provide the same scores to mark equal preferences over different translations, multiple translations can obtain the highest quality.

\begin{figure*}[t]
 \centering
\begin{subfigure}{0.45\textwidth}
  \centering
  \includegraphics[width=0.98\linewidth]{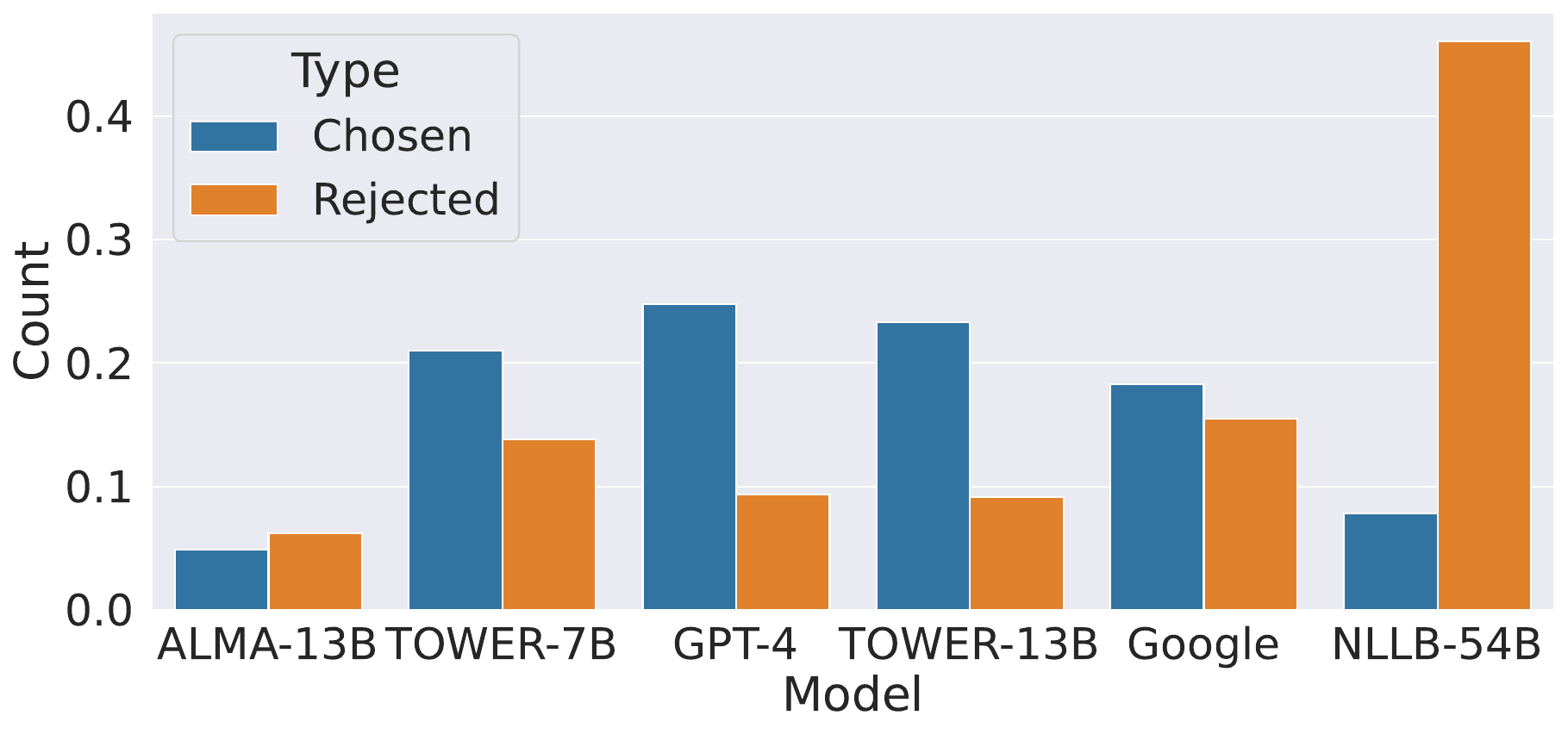}
\end{subfigure}%
% \hspace{0.3cm}
\begin{subfigure}{0.50\textwidth}
  \centering
  \includegraphics[width=0.98\linewidth]{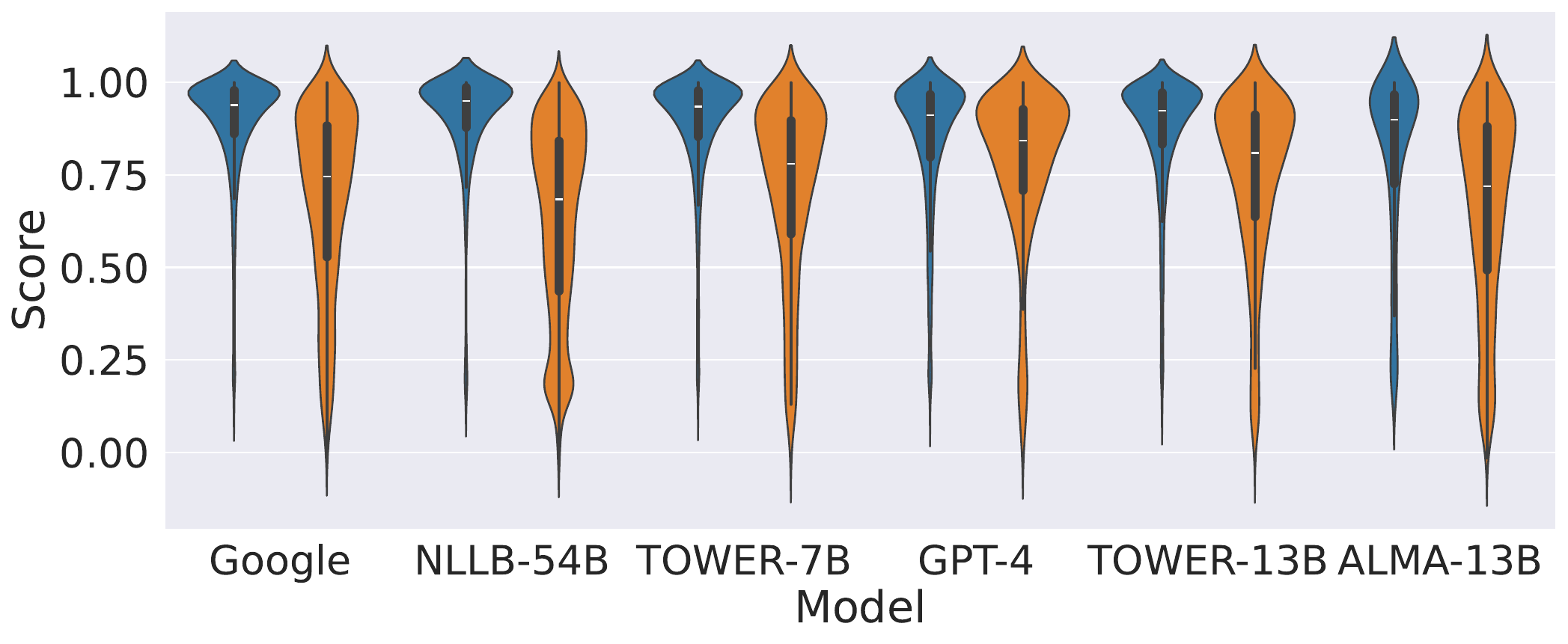}
\end{subfigure}% 
    \caption{Distribution of counts and scores for the chosen ($y_+$) and rejected ($y_-$) hypotheses across models.
    % \nllb unevenly contributes the most to $y_-$, with $y_+$ uniformly distributed between other alternatives. 
    % \andre{these images look a bit blurred, can we use a vector format instead of PNG?}
    % \nuno{we can make these figures more beautiful :); I'll follow up with you} 
    }
      \label{fig:dist_bw}
\end{figure*}

\subsubsection{Findings}\label{subsec:findings}

Our main results are summarized in Table~\ref{tab:automaticmt}.
The correlation between human judgments and metric scores on these high-quality translations is rather low, suggesting a limited ability to model human preferences between multiple translations for the same source. 
% \citep{agrawal2024automatic} \sa{maybe we should add the self-cite in later version?}
\textsc{xComet-xl+xxl}, an ensemble of \textsc{xComet-xl} and \textsc{xComet-xxl}, achieves the best Spearman and \textsc{Precision@1} across the board, even outperforming reference-based metric \textsc{Comet} \cite{rei-etal-2020-comet} on this task. Hence, we use this metric to induce preference judgments in our dataset in \S\ref{sec:data}. Designing metrics that accurately reflect these quality preferences remains an open challenge. The dataset collected in our study can potentially be used to benchmark new metrics, which we leave for future work.

\section{\toweraligndata Dataset} \label{sec:data}

Building on the findings from \S\ref{sec:human_metric}, we create our preference dataset using \textsc{xComet-xl+xxl}. We discuss the choice of the text and models in \S\ref{subsec:dataandmodel}, followed by the method for inducing and selecting preference pairs from the dataset in \S\ref{subsec:createpref}. 

\subsection{Data and MT Systems}\label{subsec:dataandmodel}
We collect source segments from \textsc{RedPajama} \cite{together2023redpajama} for English, German, French, Spanish, and Italian, and use \textsc{mc4} \cite{2019t5} for the remaining languages: Portuguese, Russian, Chinese, French, and Korean. 
Approximately 1000 segments published after July 2022 were extracted and filtered for each language using the perplexity score available in the original \textsc{RedPajama} and \textsc{mc4} collections. The perplexity thresholds vary across languages and were defined after manual checks on the filtered segments, avoiding non-fluent segments with repetitive patterns such as sequences of numbers, non-alphanumeric characters, and repeated words, among others.

We generate translation outputs using greedy decoding from six diverse models varying in architecture (encoder-decoder and decoder-only), model sizes (7B, 13B, 54B), and output quality (see Figure~\ref{fig:dist_bw}).\footnote{We did not explore alternative decoding strategies to make it easier to scale the dataset if necessary given the large sizes of the translation models.} Specifically, we use 1) \nllb \cite{costa2022no}, 2) \alma \cite{xu2023paradigm}, 3) \gpt, 4) \google, and 5) \textsc{Tower} models (\towerlarge and \towersmall). A detailed description of MT systems is provided in the Appendix~\ref{sec:model_des}. We generate translations using all models for all directions \textsc{en} $\Leftrightarrow$ \{\textsc{de, fr, pt, nl, ko, zh, ru, es, it}\}, with two exceptions. For \alma, we only generate outputs for supported language pairs (\textsc{en} $\Leftrightarrow$ \{\textsc{de, zh, ru}\}) and discard translations for \{\textsc{zh, ko}\} $\rightarrow$ \textsc{en} from \nllb due to inferior quality and frequent hallucinations.  

% \andre{haven't some of these models been introduced already? should we describe them in a single place?} \sa{I moved everything to appendix}

\subsection{Creating Preferences}\label{subsec:createpref}

For each source sentence $x$, we have up to six translation options $\{y_j\}_{j=1}^6$. Our goal is to get preference triples of source ($x$), a  preferred/chosen hypothesis ($y_{+}$), and a less preferred/rejected hypothesis ($y_{-}$). We use an automatic quality estimation metric $\mathcal{M}$ to create this dataset of preference triples $\mathcal{D} = \{ (x,y_{+}, y_{-})\}$ and resort to a simple criterion that obtains the maximum discrepancy under $\mathcal{M}$. We first measure the translation quality scores for each pair $(x, y_j)$, resulting in the scores $s = \{s_j\}_{j=1}^6$. We then select the best and the worst translation hypotheses from the ranked list induced by the scores, $s$, \textit{i.e.} $y_{+} = y_{\arg\max_j(s_j)}$ and $y_- = y_{\arg\min_j(s_j)}$. This results in a unique preference triplet for each source sentence. 
% \andre{one thing that our paper leaves open is whether it's better to convert metric scores (which these metrics already provide) to pairwise preferences and back to reward scores, vs just using the original metric scores as the reward model} \sa{that is true, maybe we can leave a footnote in the results discussion or add as a limitation?}

\section{Experimental Settings}

We use the \toweraligndata dataset\footnote{\url{https://huggingface.co/datasets/sardinelab/MT-pref}} to align MT models with translation preferences (\S\ref{sec:data}) and compare several preference learning methods detailed in \S\ref{sec:pref_methods}.  
% \andre{point also to the background which cover the several alignment algorithms we experiment with, we need to better connect these sections}. \sa{Added}

\paragraph{Training Data} 
The \toweraligndata dataset contains 18k instances with approximately 1k examples for each translation direction. 
The counts of the chosen and the rejected hypotheses from each model and the distribution of metric scores are shown in Fig.~\ref{fig:dist_bw}. The \nllb model accounts for most of the rejected hypotheses ($\sim$46\%), whereas the chosen hypotheses are more equally distributed across the \gpt, \google, and the \textsc{Tower} models, illustrating consistent and higher-quality translations generated by these models.

%for placing

\begin{table*}[!t]
    \centering
   \setlength{\tabcolsep}{2pt}
   \renewcommand{\arraystretch}{1.1}
     \resizebox{\linewidth}{!}{
    \begin{tabular}{lccccccccccc}
    % \rowcolor{gray!10}
    \toprule
    &  \multicolumn{5}{c}{\textbf{\textsc{en-xx}}} & \multicolumn{5}{c}{\textbf{\textsc{xx-en}}} & \\
    \cline{2-5} \cline{7-11}
    \addlinespace[0.1cm]
    \multirow{-2}{*}{\textbf{\textsc{Model}}} & \multicolumn{1}{c}{\textsc{chrF $\uparrow$}} & \multicolumn{1}{c}{\textsc{Comet$\uparrow$}} & \multicolumn{1}{c}{\textsc{xComet $\uparrow$}}& \multicolumn{1}{c}{\textsc{MetricX$\downarrow$}} &   & \multicolumn{1}{c}{\textsc{chrF $\uparrow$}} & \multicolumn{1}{c}{\textsc{Comet$\uparrow$}} & \multicolumn{1}{c}{\textsc{xComet $\uparrow$}}& \multicolumn{1}{c}{\textsc{MetricX$\downarrow$}} &   & \multirow{-2}{*}{\textbf{\textsc{ \% Acc. }}}  \\ 
    \midrule
\towersmall   & 52.25\fifthcluster  & 84.32\fifthcluster & 85.32\fourthcluster & 1.78\thirdcluster  && 58.87\fourthcluster & 82.77\fourthcluster & 88.77\fifthcluster & 2.20\secondcluster && 53.25 \\

~\quad + SFT  & \textbf{53.29}\fourthcluster & 84.26\fifthcluster& 85.11\fourthcluster  & 1.92\fourthcluster  & & 59.30\thirdcluster & 82.79\fourthcluster & 89.16\fourthcluster & 2.29\secondcluster && 58.50 \\ 

~\quad + DPO\textsubscript{sft} & 53.27\fourthcluster & 84.85\fourthcluster & 85.63\fourthcluster & 1.73\thirdcluster  &&  \textbf{59.86}\thirdcluster & \textbf{83.18}\thirdcluster & 89.56\secondcluster& 2.13\secondcluster  &&  59.25 \\ 

~\quad +  DPO\textsubscript{base} & 49.90\sixthcluster & 84.64\fifthcluster & 86.14\thirdcluster  & 1.44\secondcluster  &&  58.34\fourthcluster & 83.05\fourthcluster  & \textbf{89.73}\firstcluster & 1.87\firstcluster && 59.50 \\ 
 
   ~\quad + DPO\textsubscript{base}+SFT & 52.42\fifthcluster & 84.99\fourthcluster & 86.37\thirdcluster & 1.58\secondcluster &&  59.43\thirdcluster & 83.16\thirdcluster & 89.60\firstcluster& 2.03\secondcluster  && 58.25  \\  
  % \addlinespace[0.1cm]

~\quad +  CPO  & 52.95\fourthcluster & \textbf{85.05}\fourthcluster & \textbf{86.43}\thirdcluster & 1.59\secondcluster &&  59.62\thirdcluster & 83.14\thirdcluster & 89.70\firstcluster& 2.04\secondcluster && \textbf{59.50}\\

\addlinespace[0.1cm]
\hdashline
\addlinespace[0.1cm]

\towerlarge & 54.15\thirdcluster & 85.17\fourthcluster & 86.55\thirdcluster & 1.57\secondcluster  && 59.86\thirdcluster & 83.18\thirdcluster & 89.33\thirdcluster & 2.11\secondcluster  && 59.50   \\ 

~\quad + SFT  & 54.87\thirdcluster & 84.96\fourthcluster  & 86.12\thirdcluster & 1.72\thirdcluster & &  60.26\secondcluster & 83.25\secondcluster & 89.65\firstcluster &  2.16\secondcluster  && 57.25 \\

~\quad + CPO& \textbf{54.45}\thirdcluster & \textbf{85.59}\thirdcluster & \textbf{87.22}\secondcluster  & {1.43}\secondcluster  &&  \textbf{60.55}\secondcluster & \textbf{83.49}\secondcluster & \textbf{89.98}\firstcluster & \textbf{1.99}\secondcluster &&  \textbf{60.25} \\

\addlinespace[0.1cm]
\hdashline
\addlinespace[0.1cm]

\almar & 47.57\seventhcluster & 84.95\fifthcluster  & 87.27\secondcluster  & \textbf{1.30}\firstcluster &&  58.79\fourthcluster & 83.12\fourthcluster  & 89.43\secondcluster & 2.07\secondcluster && 50.00  \\

GPT-3.5 & 56.38\secondcluster & 85.56\thirdcluster & 86.92\thirdcluster &1.54\secondcluster  &&  60.92\secondcluster & 83.48\secondcluster & 90.00\firstcluster & 1.99\secondcluster &&  -  \\

\gpt & 56.94\secondcluster & 86.01\secondcluster & 87.43\secondcluster & 1.48\secondcluster  &&  61.33\secondcluster & 83.69\secondcluster & \textbf{90.34}\firstcluster& 1.82\firstcluster  && -  \\

\google & \textbf{60.43}\firstcluster & \textbf{86.44}\firstcluster  & \textbf{87.53}\firstcluster & 1.55\secondcluster  &&  \textbf{62.05}\firstcluster & \textbf{84.07}\firstcluster & 89.83\firstcluster & 2.16\secondcluster && - \\ 
\bottomrule
    \end{tabular}
    }
    \caption{Comparing PO methods on WMT23:
   Both CPO and DPO\textsubscript{base}+SFT result in significant (p=0.05) improvement in translation quality, closing the gap with \towerlarge.
    } 
    \label{tab:mainmt_avg}
\end{table*}

% \begin{figure*}
%     \centering
%     \includegraphics[width=0.39\linewidth, trim={0 0 22cm 0},clip]{figs/chosen_logps.pdf}
%    \includegraphics[width=0.59\linewidth]{figs/rejected_logps.pdf}
%     \caption{ Log probabilities for chosen (left) and rejected (right) translations during training for several PO methods: DPO\textsubscript{base} reduces the log probability for both chosen and rejected responses.   
%     }  \label{fig:loss}
% \end{figure*}

\paragraph{Evaluation}
We evaluate finetuned models on the WMT23 test set (\textsc{en} $\leftrightarrow$ \{\textsc{de, ru, zh}\}) and the FLORES dev-test set (\textsc{en} $\leftrightarrow$ \textsc{de, ru, zh, es, fr, pt, nl, it, ko})  using \textsc{Tower-Eval} \cite{alves2024tower}.\footnote{\url{https://github.com/deep-spin/tower-eval} 
% \nuno{I would save the extra line by removing the \texttt{/tree/main} bit}
} We report system-level translation quality using \textsc{chrF} \cite{popovic-2015-chrf}, \textsc{Comet}, \textsc{MetricX-XXL} \cite{juraska-etal-2023-metricx}, and \textsc{xComet-XL}. We cluster system performance using the Wilcoxon rank-sum test ($p < 0.05$) with \textsc{Comet} as the primary metric. Rank ranges, denoted by $[l+1, n-w]$, indicate the number of systems a particular system underperforms or outperforms,  where $l$ represents the number of losses, $n$ is the total number of systems, and $w$ is the number of systems that the system in question significantly outperforms \cite{kocmi-etal-2023-findings}. We compare the models' accuracy (\textsc{\% Acc.}) for selecting the best-over-worst hypothesis with the model's likelihood on the human preferences (\S\ref{sec:human_metric}) after finetuning on \toweraligndata.

\paragraph{Model Configurations}

 We finetune \towersmall using preference optimization methods detailed in \S\ref{sec:pref_methods} with the following configurations:
 \begin{itemize}
    \item SFT: a baseline model supervised finetuned on the chosen or the most preferred response.
    \item DPO\textsubscript{sft}: model trained with $\pi_\text{ref}$=SFT in Eq.~\ref{eq:dpo}.  
    \item DPO\textsubscript{base}: base model directly finetuned with DPO, \textit{i.e.} $\pi_\text{ref}$=\towersmall.
    \item DPO\textsubscript{base}+SFT: base model finetuned with a combination of DPO and SFT regularization, \textit{i.e.} $\mathcal{L}_{\text{DPO}} (x, y_\pm; \pi_\theta, \pi_\text{ref}) +  \lambda \mathcal{L}_{\text{NLL}}  (x, y_+; \theta)$.
    \item CPO: model finetuned with the objective in Eq.~\ref{eq:cpo}.
 \end{itemize}
 
We also compare the aligned models against \towerlarge (with SFT and CPO variants), \gpt, \almar\footnote{The translation outputs are directly taken from \url{https://github.com/fe1ixxu/ALMA/tree/master/outputs/wmt23_outputs/ALMA-13B-R}} and \google models. All training details are provided in Appendix~\ref{sec:hyper}. 

\section{Results}\label{sec:results}

\begin{table*}[!t]
    \centering
     \resizebox{\linewidth}{!}{
    \begin{tabular}{lccccccccc}
    % \rowcolor{gray!10}
    \toprule
    & \multicolumn{4}{c}{\textbf{\textsc{EN-XX}}} && \multicolumn{4}{c}{\textbf{\textsc{XX-EN}}}\\
    
     % \rowcolor{gray!10}
     \cline{2-5} \cline{7-10}
     \addlinespace[0.1cm]
    \multirow{-2}{*}{\textbf{\textsc{Dataset}}} & \multicolumn{1}{c}{\textsc{chrF $\uparrow$}} & \multicolumn{1}{c}{\textsc{Comet$\uparrow$}} & \multicolumn{1}{c}{\textsc{xComet $\uparrow$}}& \multicolumn{1}{c}{\textsc{MetricX$\downarrow$}} && 
    % \multicolumn{1}{c}{\textsc{Rank}} &
    \multicolumn{1}{c}{\textsc{chrF $\uparrow$}} & \multicolumn{1}{c}{\textsc{Comet$\uparrow$}} & \multicolumn{1}{c}{\textsc{xComet $\uparrow$}}& \multicolumn{1}{c}{\textsc{MetricX$\downarrow$}}  
    % \multicolumn{1}{c}{\textsc{Rank}} 
    \\ 
    \midrule
\towersmall & 56.14 & 88.51 & 93.01 & 1.05 && 64.08 & 88.28 & 96.20 & 1.18 \\ 
~\quad + CPO  & \textbf{56.70} & \textbf{88.81} & \textbf{93.71} & \textbf{0.96} &&  \textbf{64.21} & \textbf{88.32} & \textbf{96.56} & \textbf{1.15} \\ 
\addlinespace[0.1cm]
\hdashline
\addlinespace[0.1cm]
\towerlarge & 57.17 & 88.89 & 93.85 & 0.93&& 64.80 & 88.50 & 96.44& 1.12 \\ 
~\quad + CPO  & \textbf{57.79} & \textbf{89.15} & \textbf{94.30} & \textbf{0.87} &&  \textbf{64.90} & \textbf{88.51} & \textbf{96.71} &  \textbf{1.10}  \\
\bottomrule
    \end{tabular}
    }
    \caption{CPO finetuning using \toweraligndata improves translation quality for \textsc{Tower} models on FLORES.  
    % \sa{Update table for camera ready}
    }
    \label{tab:avg_flores}
\end{table*}

% \andre{maybe merge with previous section?} 
We first present the results of comparing several PO methods (\S\ref{sec:pref_methods}) in Table~\ref{tab:mainmt_avg} on the WMT23 and FLORES datasets. Scores are averaged for from-English (\textsc{en-xx}) and to-English (\textsc{xx-en}) translation directions. Results for individual language pairs are shown in Appendix~\ref{sec:by_lp}. 
We then compare preference learning on \toweraligndata against an existing preference dataset (\S\ref{subsec:compare}), followed by an ablation on the impact of the dataset size on the final translation quality (\S\ref{subsec:size}). 

\subsection{Comparing PO Algorithms}

% \andre{point to the background section to remind people what these methods are, we need to better connect these sections} \sa{updated previous section detailing the algorithmms}

\paragraph{SFT results in limited translation quality gains.} SFT 
on the \textit{chosen} response from the \toweraligndata dataset improves \textsc{chrF} over \towersmall 
% \andre{improves over what? what is the chosen response? this needs to be better explained} \sa{updated \S\ref{subsec:createpref} to describe chosen}
on \textsc{en-xx} (+1.04) and \textsc{xx-en} (+0.43) translation directions, with no significant difference in \textsc{Comet} and \textsc{xComet-XL} in \textsc{en-xx} direction. 
However, we observe a large gain (+5.25\%) in \textsc{\% Acc.}, suggesting that the model does acquire some ability to distinguish high-quality translations even when trained with best translations only. 

% \nuno{Have you tested SFT on top of TowerBase with TowerBlocks + Preferred Data?} \sa{no i haven't but that's a good suggestion, training one}

\begin{figure}
    \centering
    \includegraphics[width=\linewidth]{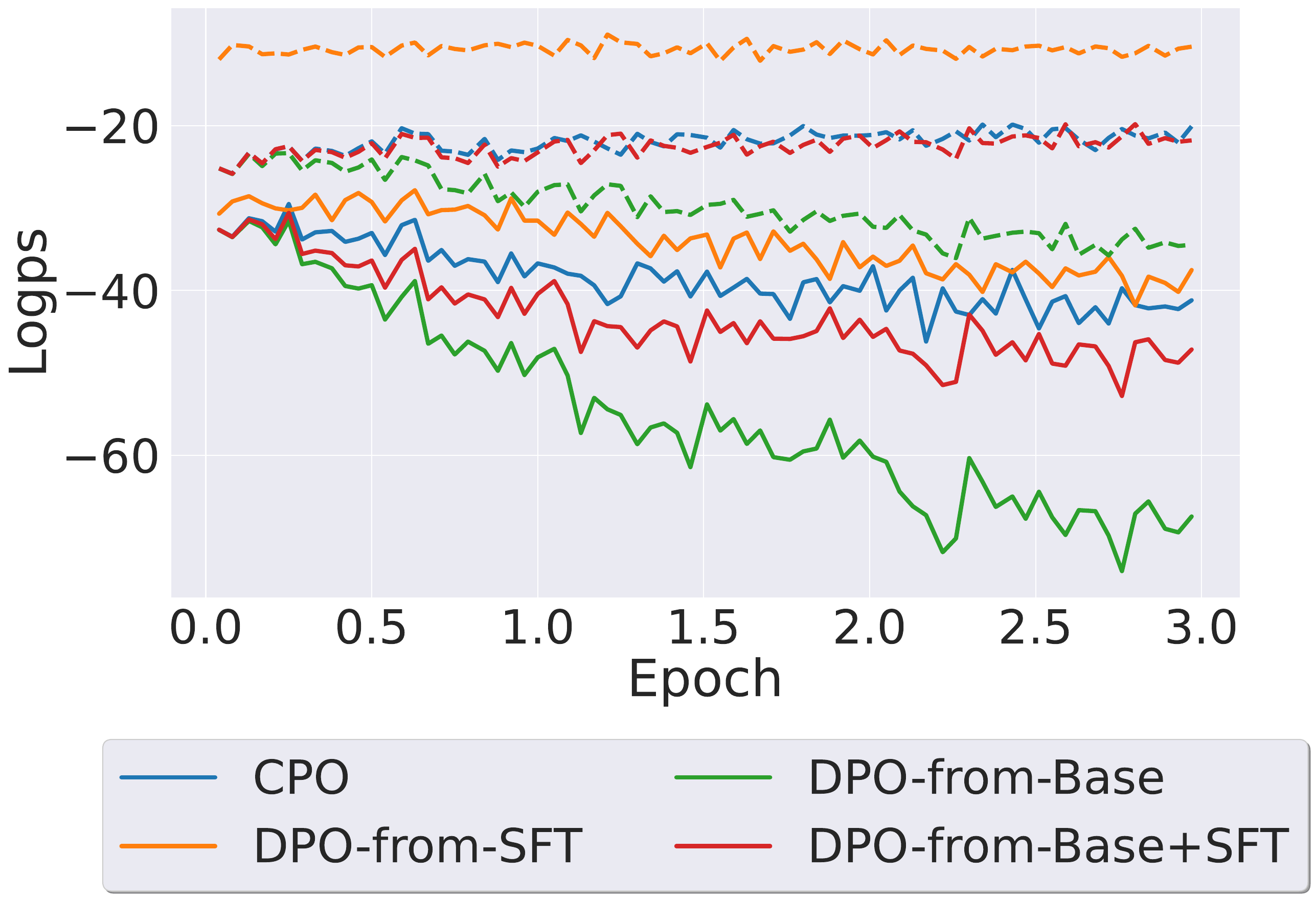}
    \caption{ Log probabilities for chosen (\rule[0.5ex]{0.2em}{0.4pt}\hspace{0.2em}\rule[0.5ex]{0.2em}{0.4pt}\hspace{0.2em}\rule[0.5ex]{0.2em}{0.4pt}) and rejected (\rule[0.5ex]{1em}{0.4pt}) hypotheses during training across PO methods: DPO\textsubscript{base} reduces the likelihood for both chosen and rejected responses, resulting in reduced output quality.  
    }  \label{fig:loss}
\end{figure}

\paragraph{Preference learning improves translation quality.} Most PO methods improve \textsc{Comet} and \textsc{xComet-XL}  as well as \textsc{\% Acc.} over \towersmall in both directions, showing that aligning LLMs with preferences benefits MT. The translation quality gap between \towersmall and \towerlarge by COMET is reduced significantly.
% to 0.12 and 0.04 \textsc{Comet} points for \textsc{en-xx} and \textsc{xx-en} language directions.
Optimizing \towerlarge on \toweraligndata with CPO further improves translation quality reaching comparable quality to \textsc{GPT-3.5} for \textsc{en-xx} and \textsc{xx-en} directions respectively. This illustrates that finetuning on \toweraligndata can improve translation quality even for larger models.

\begin{table*}[!t]
    \centering
    \setlength{\tabcolsep}{3pt}
    \resizebox{\linewidth}{!}{
    \begin{tabular}{llcccccccccc}
    % \rowcolor{gray!10}
    \toprule
    &  &  &  \multicolumn{4}{c}{\textbf{\textsc{EN-XX}}} && \multicolumn{4}{c}{\textbf{\textsc{XX-EN}}}  \\
    \cline{4-7} \cline{9-12}
     % \rowcolor{gray!10}
     \addlinespace[0.1cm]
    \multirow{-2}{*}{\textbf{\textsc{Dataset}}} &  \multirow{-2}{*}{\textbf{\textsc{Metric}}} & \multirow{-2}{*}{\textbf{\textsc{N}}} & \multicolumn{1}{c}{\textsc{chrF $\uparrow$}} & \multicolumn{1}{c}{\textsc{Comet $\uparrow$}} & \multicolumn{1}{c}{\textsc{xComet $\uparrow$}} & \multicolumn{1}{c}{\textsc{MetricX$\downarrow$}}  && \multicolumn{1}{c}{\textsc{chrF $\uparrow$}} & \multicolumn{1}{c}{\textsc{Comet $\uparrow$}} & \multicolumn{1}{c}{\textsc{xComet $\uparrow$}}  & \multicolumn{1}{c}{\textsc{MetricX$\downarrow$}} \\ 
    \midrule
\addlinespace[0.2cm]
\textsc{Tower-7b} & - & - & 52.25 & 84.32 & 85.32 &  1.78 && 58.87 & 82.77 & 88.77 & 2.20 \\
\addlinespace[0.1cm]
\hdashline
\addlinespace[0.1cm]
\multirow{3}{*}{\toweraligndata} & \multirow{2}{*}{\textsc{xComet-xl+xxl}} & 18k & 52.95 & \textbf{85.05} & \textbf{86.43} & 1.59 &&   59.62 & 83.14 & \textbf{89.70}   & 2.04\\ 
 & & 6k &  \textbf{52.98} & 84.81 & 85.98  & 1.66  &&  59.63 & 83.09 & 89.46 & 2.08\\ 
 & \textsc{xComet+Kiwi--xxl} & 18k & 52.87 & 84.86 & 85.90 & 1.66  && \textbf{59.86} & \textbf{83.15} & 89.54 & 2.07\\ 
\addlinespace[0.1cm]
\hdashline
\addlinespace[0.1cm]
\multirow{2}{*}{\textsc{ALMA-R}} & \multirow{2}{*}{\textsc{xComet+Kiwi-xxl}} &  14k & 49.87 & 84.89 & 86.35 & \textbf{1.42} && 59.63 & 83.24 & 89.47 & \textbf{2.02}\\ 
& & 6k & 51.02 & 84.76 & 85.90  & 1.52 && 59.72 & \textbf{83.15} & 89.33  & 2.07 \\ 

% \addlinespace[0.1cm]
% \toweraligndata & GPT-4 & 6k & 53.18 & 84.65 & 85.33 && 60.01 & 83.11 & 89.32 \\

\addlinespace[0.1cm]
\hdashline
\addlinespace[0.1cm]
\textsc{Tower-13b} & & & 54.15 & 85.17 & 86.55 & 1.57  && 59.86 & 83.18 & 89.33  & 2.11\\ 
\bottomrule
    \end{tabular}
    }
    \caption{CPO finetuning on \textsc{ALMA-R-Pref} and \toweraligndata variants:  Preferences induced via \textsc{xComet-xl+xxl} on all examples gives the best overall results. 
    } \label{tab:compare_data_metric}
    % \vspace{-0.3cm}
\end{table*}

\paragraph{SFT is necessary to obtain translation quality improvements using DPO.} Comparing different variants of DPO (DPO\textsubscript{sft}, DPO\textsubscript{base} and DPO\textsubscript{base}+SFT), we find that either the SFT phase or the SFT regularization is necessary to obtain significant COMET improvements. 
This also aligns with findings from \citet{tunstall2023zephyr} who show that learning from chat preference datasets fails when skipping the initial SFT stage. 
Interestingly, DPO\textsubscript{base} attains the highest \textsc{\% Acc.} scores among variants, showing an improved ability to discern but not necessarily generate high-quality translations. We find that as suggested by \citep{pal2024smaug}, it is indeed because DPO\textsubscript{base} increases the relative probability between the two classes by decreasing the model's likelihood for both \textit{chosen} and \textit{rejected} translations (see Fig.~\ref{fig:loss}).   

% \begin{figure}
%     \centering
%     \includegraphics[width=\linewidth]{figs/logps.pdf}
%     \caption{\sa{WIP}Log probabilities for PO methods during training.   
%     }  \label{fig:loss}
% \end{figure}

\paragraph{Results on FLORES} We report the results of aligning \towersmall with CPO on \textsc{Flores} in Table~\ref{tab:avg_flores}. On average, the translation quality of the base models, \towersmall and \towerlarge, improves with alignment tuning across the board according to all metrics, with  \towersmall reaching close \textsc{Comet}, \textsc{xComet-xl} and \textsc{MetricX-xxl} scores to \towerlarge, despite being 2x smaller. 

In gist, we show that CPO results in the best-aligned \towersmall, matching translation quality with \towerlarge on both WMT23 and FLORES benchmarks. We next compare preference optimization using CPO on \toweraligndata against existing preference datasets.

\subsection{\toweraligndata Vs. \textsc{ALMA-R-Pref}} \label{subsec:compare}

% \sa{explain xcomet-kiwi results, what do we loose by using a worse metric?}

% \sa{Optionally add comparison to MAPLE if and when we get the data}

% \sa{changed back to use as explore didn't sound right}
\citet{xu2024contrastive} use the FLORES-200 development and test datasets to create a preference dataset, \textsc{ALMA-R-Pref}. For each source sentence in the corpus, they take the human-written reference, and outputs from \almar and \gpt models, and induce preferences using an ensemble of \textsc{xComet}-\textsc{xxl} and \textsc{CometKiwi}-\textsc{xxl} metrics. We note that this metric ensemble attains similar or lower correlation scores compared to the best individual metrics on both language pairs as shown in Table~\ref{tab:automaticmt}. We compare the translation quality of the resulting models when aligned with \toweraligndata and \textsc{ALMA-R-Pref} preference datasets in Table~\ref{tab:compare_data_metric}.
% \footnote{We do not compare with MAPLE \cite{Zhu2024} due to lack of open access to this dataset.}

Training on \textsc{ALMA-R-Pref} preference dataset improves neural metrics but significantly hurts \textsc{chrF} compared to the base model, \towersmall.\footnote{A difference of 2.4 \textsc{chrF} points is considered significant with 87\% accuracy \cite{kocmi2024navigating}.} Our analysis shows that finetuning on the \textsc{ALMA-R-Pref} dataset increases the output length significantly. This could be due to the inherent bias in the dataset where the \textit{chosen} responses, typically by \gpt (45\%), are on average longer than the \textit{rejected} responses.\footnote{The difference in the length of \textit{chosen} and \textit{rejected} translations in the training dataset is also significant according to an independent t-test with a p-value of $0.01$.} This has important implications for the creation and modeling of preferences \---\ when a model is too frequently ``preferred'' in a dataset, it can lead to the distillation of that model's characteristics and it is unclear to what extent humans prefer these distilled features.

\towersmall finetuned on equal-sized \textsc{ALMA-R-Pref} and \toweraligndata datasets score close on neural metrics (\textsc{comet} and \textsc{xcomet}), with a difference of 1.96 points on \textsc{chrF}. As our preference dataset considers outputs from multiple models with diverse styles, we do not distill any such model-specific biases. Furthermore, aligning on preferences induced via \textsc{xComet-xl+xxl} yields slightly better \textsc{Comet} score on \textsc{en-xx} direction over preferences with \textsc{xComet+Kiwi-xxl}, further validating the importance of inducing preferences using metrics guided by human knowledge.

\subsection{Impact of the Size of Preference Datasets} \label{subsec:size}

\begin{figure}
    \centering
    \includegraphics[width=0.90\linewidth]{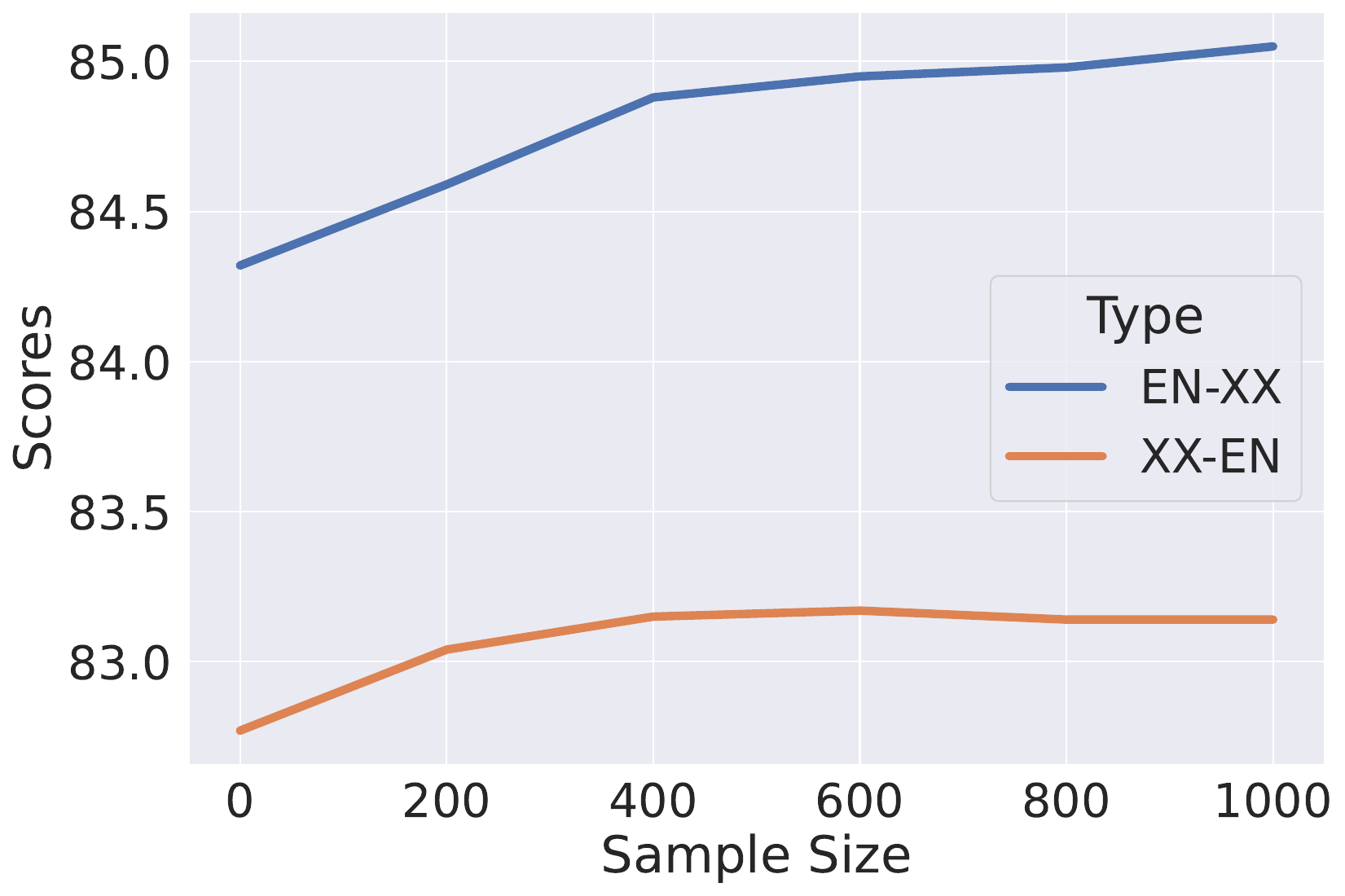}
    \caption{ COMET with varying size of the preference dataset: \textsc{en-xx} continues to benefit from more samples.
    % \andre{this figure is blurry, can we get a plot in vector format instead of a PNG, or at least increase resolution?}
    }  \label{fig:size}
\end{figure}

One advantage of our approach is that we can scale the size of preference datasets as necessary as preferences are induced using an automatic QE metric. To understand whether this is indeed beneficial, we conduct an ablation where we vary the number of unique source samples per language pairs as: \{200, 400, 600, 800, 1000\} and align \towersmall on the resulting preference dataset using CPO. Fig.~\ref{fig:size} shows the results: while the improvement in quality for \textsc{xx-en} plateaus with just 400 samples per language direction, \textsc{Comet} continues to improve for \textsc{en-xx} suggesting that adding more data might benefit translations from English to other language pairs. This aligns with the fact that the model is exposed to relatively fewer non-English texts during pretraining and hence benefits more from any additional dataset on these languages.

\section{Related Work}

\paragraph{LLMs for MT}
Earlier works exploring LLMs to perform MT study prompting techniques to generate translations \cite{hendy2023good, zhang2023prompting, vilar-etal-2023-prompting} with research focusing on selecting high-quality and relevant examples as demonstrations to incorporating external knowledge mimicking human-like translation strategies \cite{he2023exploring}. More recently, several works have proposed finetuning LLMs to improve the translation quality \cite{zhang-etal-2023-machine, alves-etal-2023-steering}, resulting in specialized models that attain competitive performance to state-of-the-art production level translation systems \cite{xu2023paradigm, alves2024tower}. 
Across all methods, the quality of the data used for training is paramount to the finetuning methods. 
% success of the translation capabilities of the LLMs. 
Therefore, in this work, we focus on curating a high-quality translation preference dataset using metrics that closely reflect true human translation preferences and outputs generated from a diverse set of high-quality MT systems.

\paragraph{Quality Feedback for MT}
Using feedback from automatic metrics for MT 
 or human quality assessment has been an active area of research through the past decade. 
 % \antonio{These are a lot of citations and it's hard to tell which ones follow in each category. can we cite them after training, decoding, and directly?}
 This quality signal is either utilized during training \cite{shen-etal-2016-minimum, wieting-etal-2019-beyond, yang2023direct, he2024improving, gulcehre2023reinforced, nguyen-etal-2017-reinforcement, kreutzer-etal-2018-neural, kreutzer-etal-2020-correct} or decoding \cite{freitag-etal-2022-high, fernandes-etal-2022-quality, farinhas-etal-2023-empirical} or for modeling translation preferences in the dataset directly \cite{xu2024contrastive, Zhu2024}. Similar to \citet{xu2024contrastive}, we use automatic metrics to induce preferences in the dataset but with the additional validation that the chosen metric indeed reflects human quality expectations and with translations generated from diverse MT systems.

 % \todo{any other relevant related work}
 
 % \nuno{I would either cite all papers after ''training'' and ''decoding''. or I would not cite any paper here. Seems to be easier to move Patrick's paper to the first bunch of citations.}. \sa{expand}
% \paragraph{}

\section{Conclusion}

We present \toweraligndata, a high-quality translation preference dataset, curated by combining the strengths of human evaluation and automatic metrics. The dataset includes metric-induced preferences from strong MT models across 18 language directions with new source sentences mined post-2022.  Aligning state-of-the-art decoder-only LLMs on this preference dataset using existing aligning tuning algorithms improves translation quality. Furthermore, the aligned models are also better at modeling human preferences of translation quality. 
% \nuno{hanging line} \sa{will take care of all widows once we have everything in the paper}

\section*{Limitations}
We note a few limitations of our work. We evaluate the translation quality of the finetuned models primarily using automatic metrics. While we validate that they can indeed provide a reasonable signal to differentiate quality at the system level (See Appendix~\ref{sec:system}), it requires a human evaluation to confirm whether and to what extent the aligned models match human preferences. 
Furthermore, we use existing QE metrics that can be sensitive to the domain of the datasets \cite{zouhar2024finetuned}. However, as the QE metrics continue to improve, our approach allows to substitute the preferences with that induced by a better QE metric. 
Finally, we do not handle tied preferences in translation quality and always induce a strict preference order. Incorporating neutral preferences between translations can help the model focus on attributes that truly improve quality over stylistic preferences; we leave the investigation of this phenomenon to future work. 
We note that our dataset can be used to design better QE metrics for ranking translations, inducing preferences using new criteria, and employing better optimization methods. 

\section*{Potential Risks}
Large language models may carry the potential risk of generating fluent and hallucinated content. When the users do not know the target or the source language, they might trust the generated translation without further verification \cite{martindale2018fluency}. And while our approach is driven toward making the model aware of translations of varying quality during finetuning, the coverage is limited to the supported language pairs. Users should exercise caution and seek verification from additional sources where possible when using LLMs on real-world applications.

\section*{Acknowledgments}

We thank Miguel Ramos, Miguel Faria, and Giuseppe Attanasio for their useful and constructive comments. This work was supported by the Portuguese Recovery and Resilience Plan through project C645008882-00000055 (Center for Responsible AI), by the EU’s Horizon Europe Research and Innovation Actions (UTTER, contract 101070631), by the project DECOLLAGE (ERC-2022-CoG 101088763), and by Fundação para a Ciência e Tecnologia through contract UIDB/50008/2020.

\bibliography{anthology, custom}

\appendix
\newpage
\onecolumn
\section{Annotation Guidelines and Interface} \label{sec:guidelines}

\paragraph{Task Overview} This task involves evaluating five translations of a source text and assigning a quality rating to each translation based on its overall quality and adherence to the source content. You will need to consider the accuracy, fluency, and overall quality when assessing the different translations. 

\paragraph{Annotation Scale} Each translation is evaluated on a continuous scale of 0-6 with the quality levels described as follows: 

\begin{itemize}
    \item 6: Perfect Meaning and Grammar: The meaning of the translation is completely consistent with the source and the surrounding context (if applicable). The grammar is also correct. 
    \item 4: Most Meaning Preserved and Few Grammar Mistakes: The translation retains most of the meaning of the source. It may have some grammar mistakes or minor contextual inconsistencies. 
    \item 2: Some Meaning Preserved: The translation preserves some of the meaning of the source but misses significant parts. The narrative is hard to follow due to fundamental errors. Grammar may be poor. 
    \item 0: Nonsense/No meaning preserved: Nearly all information is lost between the translation and source. Grammar is irrelevant.
\end{itemize}

\begin{figure}[t]
 \centering
\begin{subfigure}{\textwidth}
  \centering
  \includegraphics[width=\linewidth, scale=0.75]{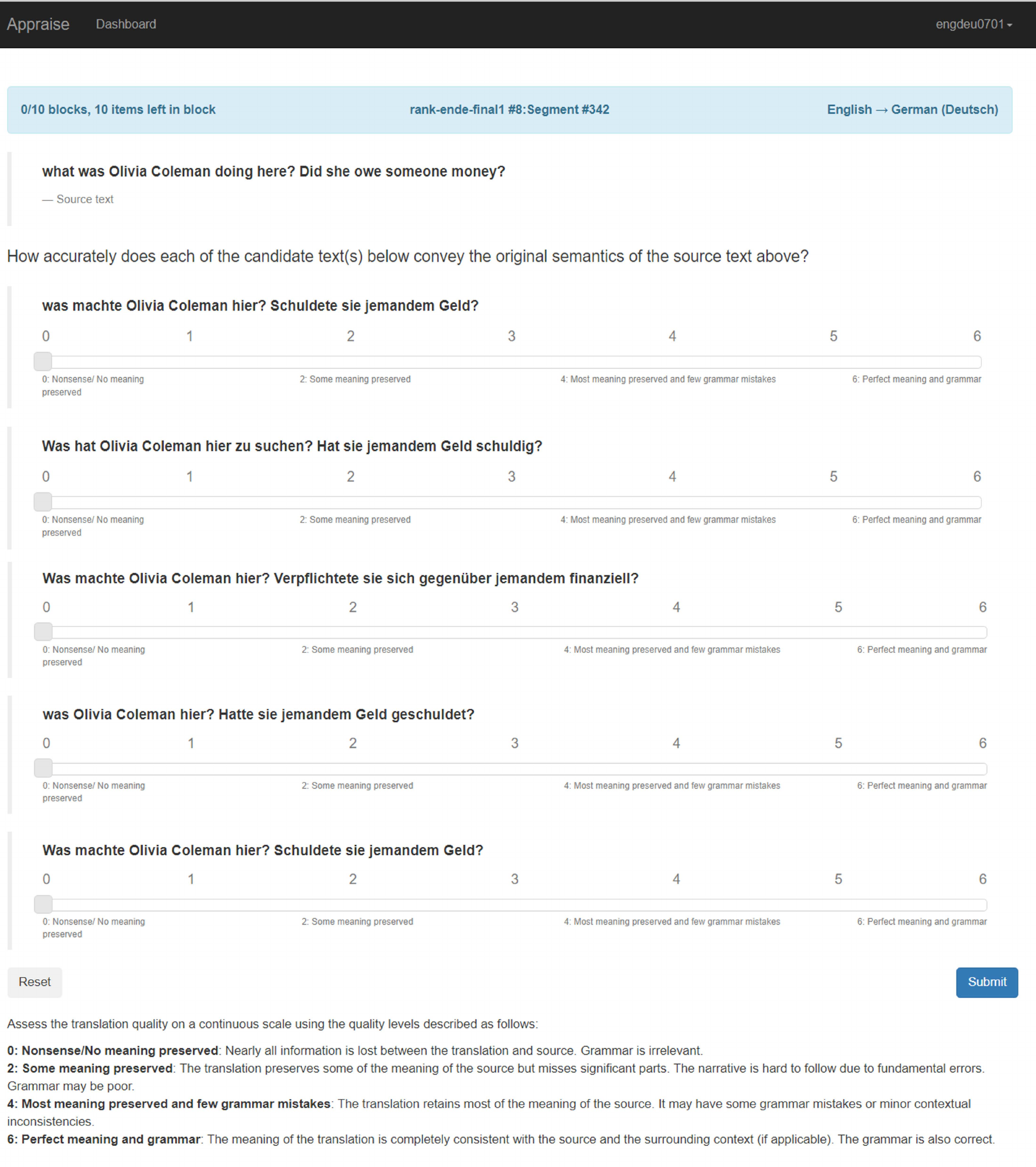}
\end{subfigure}%
    \caption{Annotation Interface.}
      \label{fig:interface}
\end{figure}

You can scroll up or down to see all the other translation outputs from the different systems. Figure~\ref{fig:interface} shows the interface when comparing and evaluating five translations. While each translation is evaluated independently, these translations can also be ranked based on the difference in their absolute scores. It is perfectly valid to give the same score to multiple translations if you believe they are of the same overall quality.

\paragraph{Other Details} We hired native speakers of Chinese and German for this task (both females) and they were compensated at \$20 per hour.

\section{MT Systems} \label{sec:model_des}

We use the following MT systems:
\begin{enumerate}
    \item \textbf{NLLB-54B} \cite{costa2022no} is a 54B encoder-decoder multilingual translation model, based on a sparsely gated Mixture of Experts (MoE) approach.  It covers 202 languages, supporting translation for many low-resource languages.
    \item \textbf{\towerlarge and \towersmall} are 13B and 7B decoder-only LLMs, trained to optimize quality on multiple tasks present in translation workflows. The model is continued pretrained from \textsc{Llama 2} \cite{touvron2023llama} checkpoints on a multilingual mixture of monolingual and parallel data, followed by finetuning on instructions relevant to translation processes. The model is instructed with the following prompt using the chat template to generate the translation output:
    \cmss{ Translate the following [source language] source text to [target language]:} \\
    \cmss{[source language]: [source sentence]} \\
    \cmss{[target language]:}
    
    \item \textbf{\textsc{ALMA-13b}} \cite{xu2023paradigm} is a 13B decoder-only model specialized for MT via continued pretaining, followed by instruction tuning on a small but high-quality parallel dataset. Unlike \textsc{TowerInstruct} models, the continued pretraining phase only explores monolingual data, and the instruction tuning is performed with an MT dataset only.
    \item \textbf{\textsc{ALMA-13b-R}}  \cite{xu2024contrastive} is a 13B decoder-only model obtained by finetuning \alma with \textsc{ALMA-R-Pref} using CPO. \\
    We use the prompt from the original paper to generate translations using ALMA models:\\
    \cmss{Translate this from [source language] to [target language]:} \\
    \cmss{[source language]: [source sentence]} \\
    \cmss{[target language]:}

    % \todo{update the below to match style of top 4}
    \item \textbf{\textsc{GPT-4}} \cite{achiam2023gpt} is prompted in a zero-shot fashion, following \citet{hendy2023good}, to generate translations using the prompt: \\
    \cmss{Translate this sentence from [source language] to [target language]:} \\
    \cmss{Source: [source sentence]} \\
    \cmss{Target:}
    \item \textbf{\google} is the basic version of the Translate API v2 accessed on \textsc{2024-03-04}.\footnote{\url{https://translation.googleapis.com/language/translate/v2}}
\end{enumerate}

\section{System-level Correlation}   \label{sec:system}

Table~\ref{tab:automaticmtsystem} shows the system-level translation quality scores assigned by reference-based metrics: \textsc{chrF}, \textsc{Comet}, and \textsc{xComet}-\textsc{xl} for all five models and their induced system-level rankings. For both directions, \textsc{Comet} results in 90\% agreement with human judgments, confirming its accuracy in rating high-quality systems and hence we use \textsc{Comet} as the primary metric for ranking different systems. 
% \andre{isn't it a bit surprising that COMET is better than more recent metrics like xCOMET-XL?} \sa{maybe due to the size and nature of the dataset?}
% 

\begin{table*}[!t]
    \centering
   \resizebox{0.98\linewidth}{!}{
    \begin{tabular}{lccccccccc}
    \rowcolor{gray!10}
   & \multicolumn{4}{c}{\textbf{\textsc{en-de}}} && \multicolumn{4}{c}{\textbf{\textsc{zh-en}}}  \\
   \rowcolor{gray!10}
   \multirow{-2}{*}{\textbf{\textsc{MODEL}}} & \textsc{chrF}  & \textsc{COMET} & \textsc{xComet}-\textsc{xl}  & \textsc{DA}&& \textsc{chrF}  & \textsc{COMET} & \textsc{xComet}-\textsc{xl}  & \textsc{DA}  \\
    \addlinespace[0.5em]
   \google & $68.83$ (1) &	$0.854$ (1)	& $0.941$ (1) &	$86.87$ (2)  && $49.40$ (1) & $0.810$ (1) & $0.884$ (1) & $79.85$ (1)\\
    \gpt & $68.50$ (2) &	$0.848$ (2) &	$0.932$ (3) &	$87.98$ (1) && $45.95$ (2) & $0.799$ (2) & $0.877$ (2) & $79.12$ (2)\\ 
 \towerlarge & $66.45$ (3) &	$0.843$ (3) &	$0.931$ (4)&	$86.53$ (3) && $45.29$ (3) & $0.794$ (3) & $0.866$ (3) & $69.12$ (3)\\
 \almar & $59.92$ (5) &$0.836$ (4)	&$0.935$ (2)	&$84.96$ (4) && $44.72$ (4) & $0.793$ (4) & $0.858$ (5) & $66.02$ (5)\\
  \towersmall & $64.61$ (4)	&$0.830$ (5) &	$0.918$ (5)	&$83.32$ (5) && $43.77$ (5) & $0.790$  (5) & $0.860$ (4) & $68.66$ (4)\\

\addlinespace[0.1cm]
\hdashline
\addlinespace[0.1cm]
\textsc{Pairwise-Acc} & \multicolumn{1}{c}{8/10} & \multicolumn{1}{c}{9/10} & \multicolumn{1}{c}{7/10} & - && \multicolumn{1}{c}{9/10} & \multicolumn{1}{c}{9/10} & \multicolumn{1}{c}{10/10} & -\\
  
    \end{tabular}
    }
    \caption{Automatic Evaluation - System Level for reference-based metrics. Ranks represent the ordering based on averaged DA scores.}
    \label{tab:automaticmtsystem}
\end{table*}

\section{Training Details} \label{sec:hyper}

\paragraph{Hyperparameters} We finetune \towersmall and \towerlarge models \cite{alves2024tower} using the TRL library \cite{vonwerra2022trl} with a batch size of 64, a maximum output length of 256, a learning rate of $5\times 10^{-7}$ and a warm-up ratio of 0.1. The model is finetuned using different preference algorithms (\S\ref{sec:pref_methods}) for 3 epochs with RMSProp optimizer \cite{hinton2012neural}. For SFT, following \cite{tunstall2023zephyr}, we finetune the base model for one epoch with a learning rate of $1\times 10^{-5}$ using Adam optimizer \cite{kingma2014adam}. We use greedy decoding to generate translation hypotheses using the aligned models. All our models are trained on two Nvidia A100 GPUs. Training takes approximately four to five hours to converge.

\section{Results by WMT23 Language Direction} \label{sec:by_lp}

We report results comparing preference optimization methods when trained with \toweraligndata on individual language pairs using \textsc{Comet}, \textsc{chrF}, \textsc{xComet-xl} and \textsc{MetricX-xxl} in Tables~\ref{tab:mt_comet_lp}, ~\ref{tab:mt_chrf_lp}, ~\ref{tab:mt_xcomet_lp} and ~\ref{tab:mt_metricX_lp} respectively.

\begin{table*}[h]
    \centering
    \resizebox{0.9\linewidth}{!}{
    \begin{tabular}{lrrrrrr}
    % \rowcolor{gray!10}
    \toprule
   \multicolumn{1}{c}{\textbf{\textsc{Model}}} & \multicolumn{1}{c}{\textsc{EN-DE}} & \multicolumn{1}{c}{\textsc{EN-ZH}} & \multicolumn{1}{c}{\textsc{EN-RU}} & \multicolumn{1}{c}{\textsc{DE-EN}}  & \multicolumn{1}{c}{\textsc{ZH-EN}} & \multicolumn{1}{c}{\textsc{RU-EN}} \\ 
   \midrule
\addlinespace[0.2cm]
\towersmall   &  83.25 &  84.98 &  84.72 &  85.25 &  80.15 &  82.90   \\ 
\addlinespace[0.1cm]
~\quad + SFT  &  83.01 &  85.47 &  84.29 &  85.25 &  80.25 &  82.86    \\ 
  \addlinespace[0.1cm]
~\quad + DPO\textsubscript{sft} &  83.83 &  85.81 &  84.91 &  85.66 &  80.72 &  83.17 \\ 
~\quad +  DPO\textsubscript{base} &  83.73 &  84.64 &  85.55 &  85.25 &  80.60 &  83.30   \\ 
   ~\quad + DPO\textsubscript{base}+SFT &  83.86 &  85.65 &  85.46 &  85.53 &  80.69 &  83.26  \\  
  \addlinespace[0.1cm]
~\quad +  CPO  &  83.92 &  85.74 &  85.49 &  85.47 &  80.79 &  83.17  \\ 
\addlinespace[0.1cm]
\hdashline
\addlinespace[0.1cm]
\towerlarge &  84.02 &  85.97 &  85.52 &  85.60 &  80.71 &  83.23   \\ 
~\quad + SFT & 83.73 & 86.00 & 85.15 & 85.62 & 81.04 & 83.08
 \\
~\quad + CPO&  84.53 &  86.32 &  85.91 &  85.72 &  81.25 &  83.49  \\
\addlinespace[0.1cm]
\hdashline
\addlinespace[0.1cm]
\almar  &  84.03 &  84.97 &  85.85 &  85.54 &  80.55 &  83.28  \\ 
\textsc{GPT-3.5} &  84.61 &  86.70 &  85.38 &  85.91 &  81.52 &  83.02  \\ 
\gpt &  84.89 &  87.08 &  86.07 &  86.17 &  81.27 &  83.63  \\ 
\google   &  84.77 &  88.09 &  86.45 &  86.24 &  82.19 &  83.78  \\ 
\bottomrule
    \end{tabular}
    }
    \caption{\textsc{Comet} ($\uparrow$) on WMT23 dataset comparing PO methods when trained with \toweraligndata.
    } 
    \label{tab:mt_comet_lp}
\end{table*}

\begin{table*}[h]
    \centering
    \resizebox{0.9\linewidth}{!}{
    \begin{tabular}{lrrrrrr}
    \toprule
   \multicolumn{1}{c}{\textbf{\textsc{Model}}} & \multicolumn{1}{c}{\textsc{EN-DE}} & \multicolumn{1}{c}{\textsc{EN-ZH}} & \multicolumn{1}{c}{\textsc{EN-RU}} & \multicolumn{1}{c}{\textsc{DE-EN}}  & \multicolumn{1}{c}{\textsc{ZH-EN}} & \multicolumn{1}{c}{\textsc{RU-EN}} \\ 
   \midrule
\towersmall    &  65.74 &  37.34 &  53.66 &  67.80 &  49.91 &  58.89  \\ 
\addlinespace[0.1cm]
~\quad + SFT   &  65.76 &  40.16 &  53.95 &  67.93 &  50.89 &  59.08  \\  
  \addlinespace[0.1cm]
~\quad + DPO\textsubscript{sft} &  66.16 &  39.56 &  54.10 &  68.57 &  51.61 &  59.38  \\ 
~\quad +  DPO\textsubscript{base} &  64.23 &  33.02 &  52.43 &  66.61 &  50.25 &  58.15  \\ 
   ~\quad + DPO\textsubscript{base}+SFT &  65.90 &  37.59 &  53.78 &  67.97 &  50.89 &  59.42  \\   
  \addlinespace[0.1cm]
~\quad +  CPO  &  66.22 &  38.68 &  53.96 &  68.25 &  51.31 &  59.30  \\
\addlinespace[0.1cm]
\hdashline
\addlinespace[0.1cm]
\towerlarge &  66.90 &  40.62 &  54.95 &  68.47 &  51.22 &  59.89  \\
~\quad + SFT & 67.39 & 41.98 & 55.25 & 68.42 & 52.38 & 59.99 \\
~\quad + CPO&  67.36 &  40.74 &  55.24 &  69.03 &  52.35 &  60.28  \\ 
\addlinespace[0.1cm]
\hdashline
\addlinespace[0.1cm]
\almar  &  60.38 &  32.14 &  50.19 &  66.30 &  51.28 &  58.79  \\ 
\textsc{GPT-3.5} &  68.38 &  45.25 &  55.50 &  69.21 &  53.78 &  59.77  \\ 
\gpt  &  69.30 &  45.67 &  55.86 &  69.91 &  53.37 &  60.70  \\ 
\google  &  69.08 &  52.99 &  59.21 &  70.28 &  55.15 &  60.72  \\ 
\bottomrule
    \end{tabular}
    }
    \caption{\textsc{chrF} ($\uparrow$) on WMT23 dataset comparing PO methods when trained with \toweraligndata.
    } 
    \label{tab:mt_chrf_lp}
\end{table*}

\begin{table*}[h]
    \centering
   \resizebox{0.9\linewidth}{!}{
    \begin{tabular}{lrrrrrr}
    \toprule
   \multicolumn{1}{c}{\textbf{\textsc{Model}}} & \multicolumn{1}{c}{\textsc{EN-DE}} & \multicolumn{1}{c}{\textsc{EN-ZH}} & \multicolumn{1}{c}{\textsc{EN-RU}} & \multicolumn{1}{c}{\textsc{DE-EN}}  & \multicolumn{1}{c}{\textsc{ZH-EN}} & \multicolumn{1}{c}{\textsc{RU-EN}} \\ 
   \midrule
\towersmall    &  84.44 &  83.77 &  87.75 &  89.07 &  85.02 &  92.23   \\ 
\addlinespace[0.1cm]
~\quad + SFT  &  84.48 &  83.67 &  87.19 &  89.24 &  85.75 &  92.48  \\  
  \addlinespace[0.1cm]
~\quad + DPO\textsubscript{sft}  &  84.98 &  84.13 &  87.78 &  89.62 &  86.22 &  92.84  \\ 
~\quad +  DPO\textsubscript{base} &  85.17 &  83.78 &  89.47 &  89.54 &  86.41 &  93.23  \\ 
   ~\quad + DPO\textsubscript{base}+SFT &  85.24 &  84.67 &  89.20 &  89.51 &  86.33 &  92.95   \\   
  \addlinespace[0.1cm]
~\quad +  CPO &  85.33 &  84.98 &  88.97 &  89.51 &  86.70 &  92.88  \\ 
\addlinespace[0.1cm]
\hdashline
\addlinespace[0.1cm]
\towerlarge &  85.42 &  85.17 &  89.05 &  89.41 &  85.81 &  92.77  \\
~\quad + SFT & 85.06 & 84.57 & 88.74 & 89.74 & 86.33 & 92.87\\
~\quad + CPO &  86.13 &  85.80 &  89.74 &  89.86 &  86.86 &  93.21  \\ 
\addlinespace[0.1cm]
\hdashline
\addlinespace[0.1cm]
\almar  &  86.09 &  84.81 &  90.91 &  89.24 &  86.14 &  92.92  \\ 
\textsc{GPT-3.5}   &  86.62 &  85.16 &  88.99 &  89.80 &  87.23 &  92.98  \\ 
\gpt  &  86.72 &  85.59 &  89.98 &  89.92 &  87.43 &  93.68  \\ 
\google  &  85.76 &  86.73 &  90.11 &  89.37 &  86.93 &  93.20  \\ 
\bottomrule
    \end{tabular}
    }
    \caption{\textsc{xComet-XL} ($\uparrow$) on WMT23 dataset comparing PO methods when trained with \toweraligndata.
    } 
    \label{tab:mt_xcomet_lp}
\end{table*}

\begin{table*}[h]
    \centering
   \resizebox{0.9\linewidth}{!}{
    \begin{tabular}{lrrrrrr}
    \toprule
   \multicolumn{1}{c}{\textbf{\textsc{Model}}} & \multicolumn{1}{c}{\textsc{EN-DE}} & \multicolumn{1}{c}{\textsc{EN-ZH}} & \multicolumn{1}{c}{\textsc{EN-RU}} & \multicolumn{1}{c}{\textsc{DE-EN}}  & \multicolumn{1}{c}{\textsc{ZH-EN}} & \multicolumn{1}{c}{\textsc{RU-EN}} \\ 
   \midrule
\towersmall    & 2.13 & 1.43 & 1.77 & 2.17 & 2.49 & 1.94 \\ 
\addlinespace[0.1cm]
~\quad + SFT  &  2.24 & 1.54 & 1.97 & 2.34 & 2.56 & 1.96\\  
  \addlinespace[0.1cm]
~\quad + DPO\textsubscript{sft}  & 1.96 & 1.43 & 1.80 & 2.14 & 2.37 & 1.89   \\ 
~\quad +  DPO\textsubscript{base} & 1.64 & 1.26 & 1.42 & 1.83 & 2.08 & 1.70   \\ 
   ~\quad + DPO\textsubscript{base}+SFT &1.90 & 1.28 & 1.55 & 2.05 & 2.21 & 1.82    \\   
  \addlinespace[0.1cm]
~\quad +  CPO &  1.93 & 1.29 & 1.56 & 2.06 & 2.20 & 1.86 \\ 
\addlinespace[0.1cm]
\hdashline
\addlinespace[0.1cm]
\towerlarge &  1.87 & 1.28 & 1.55 & 2.05 & 2.44 & 1.84 \\
~\quad + SFT & 2.00 & 1.44 & 1.73 & 2.15 & 2.43 & 1.91   \\ 
~\quad + CPO & 1.64 & 1.22 & 1.44 & 1.98 & 2.21 & 1.78\\
\addlinespace[0.1cm]
\hdashline
\addlinespace[0.1cm]
\almar  &  1.77 & 1.30 & 1.57 & 2.16 & 2.47 & 1.8  \\ 
\textsc{GPT-3.5}   & 1.71 & 1.23 & 1.49 & 1.79 & 2.02 & 1.64  \\ 
\gpt  & 1.68 & 1.32 & 1.62 & 2.05 & 2.11 & 1.82 \\ 
\google  &1.46 & 1.19 & 1.25 & 2.19 & 2.25 & 1.78 \\ 
\bottomrule
    \end{tabular}
    }
    \caption{\textsc{MetricX} ($\downarrow$) on WMT23 dataset comparing PO methods when trained with \toweraligndata.
    } 
    \label{tab:mt_metricX_lp}
\end{table*}

% \section{Training Loss} \label{sec:loss}

% \begin{figure}
%     \centering
%     \includegraphics[width=0.50\linewidth]{figs/chosen.png}
%    \includegraphics[width=0.50\linewidth]{figs/rejected.png}
%     \caption{Training Loss for different preference optimization methods.}  \label{fig:loss}
% \end{figure}

% \section{Licenses}

% \section{AI Assistants}
% We have used ChatGPT\footnote{\url{https://chat.openai.com/}} during paper writing for paraphrasing or polishing original content.

\end{document}